\pdfoutput=1
\documentclass[11pt]{article}

\usepackage[final]{acl}

\usepackage{times}
\usepackage{latexsym}

\usepackage[T1]{fontenc}
\usepackage[utf8]{inputenc}

\usepackage{microtype}

\usepackage{inconsolata}

\usepackage{graphicx}

\usepackage{scrextend}
\usepackage{colortbl}

\usepackage{times}
\usepackage{latexsym}
\usepackage{graphicx}
\usepackage{booktabs}
\usepackage{multirow}
\usepackage{arydshln}
\usepackage{comment}
\usepackage{multirow}
\usepackage{amssymb}
\usepackage{amsmath}
\usepackage{bclogo}
\usepackage[skins]{tcolorbox}
\usepackage{rotating}
\usepackage{graphicx}
\usepackage[T1]{fontenc}
\usepackage{footmisc}
\usepackage{enumitem}
\usepackage{float}

\title{Overview of PerpectiveArg2024\\ The First Shared Task on Perspective Argument Retrieval}

\author{Neele Falk\thanks{* Equal contribution.}$^{1}$,
Andreas Waldis $^{*2,3}$,
Iryna Gurevych$^{2}$\\
$^1$Institute for Natural Language Processing, University of Stuttgart, Germany\\
$^2$Ubiquitous Knowledge Processing Lab (UKP Lab) \\
Department of Computer Science and Hessian Center for AI (hessian.AI)\\
Technical University of Darmstadt\\
 $^3$Information Systems Research Lab, Lucerne University of Applied Sciences and Arts \\
\texttt{\href{http://www.ukp.tu-darmstadt.de/}{www.ukp.tu-darmstadt.de}} \hspace{0.5em} \texttt{\href{http://www.hslu.ch/}{www.hslu.ch}}\\
}
\begin{document}
\maketitle

\begin{abstract}
Argument retrieval is the task of finding relevant arguments for a given query.
While existing approaches rely solely on the semantic alignment of queries and arguments, this first shared task on perspective argument retrieval incorporates perspectives during retrieval, accounting for latent influences in argumentation.
We present a novel multilingual dataset covering demographic and socio-cultural (\textit{socio}) variables, such as age, gender, and political attitude, representing minority and majority groups in society. 
We distinguish between three scenarios to explore how retrieval systems consider explicitly (in both query and corpus) and implicitly (only in query) formulated perspectives.
This paper provides an overview of this shared task and summarizes the results of the six submitted systems.
We find substantial challenges in incorporating perspectivism, especially when aiming for personalization based solely on the text of arguments without explicitly providing \textit{socio} profiles. 
Moreover, retrieval systems tend to be biased towards the majority group but partially mitigate bias for the female gender.
While we bootstrap perspective argument retrieval, further research is essential to optimize retrieval systems to facilitate personalization and reduce polarization.\footnote{Please find evaluation code and further information on \href{https://github.com/Blubberli/perspective-argument-retrieval}{https://github.com/Blubberli/argmin2024-perspective}.}
\end{abstract}

\section{Introduction}\label{sec:introduction}

%Argumentation and the automatic processing of such arguments (argument mining) are critical in political and everyday discourse. Important challenges in the field of AM are the automated analysis of argumentative structures (recognition of stance, claim, and premise), the classification of argument quality, or the automatic summarization and extraction of arguments based on large amounts of data.

In argument retrieval, the objective is to extract arguments that match a given query, such as a question or topic. 
Existing research defines the relevance and ordering of candidate arguments differently. 
In the simplest case, arguments are extracted based on the semantic relevance of the query. 
More sophisticated methods consider the quality of the arguments, suitable counterarguments \cite{WachsmuthSS18}, or arguments that answer comparative questions \cite{bondarenko22}. 
However, incorporating individual perspectives \citep{DBLP:conf/aaai/CabitzaCB23} is crucially understudied.

\begin{figure}[t]
    \centering
    \includegraphics[width=0.48\textwidth]{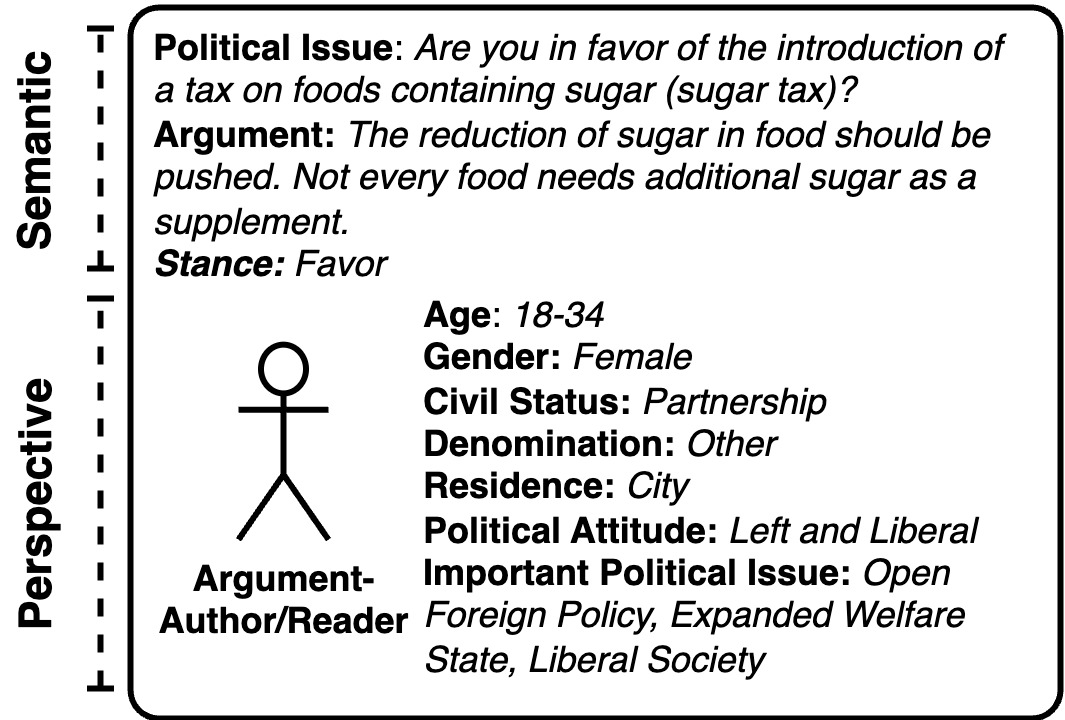}
    
         \vspace{-1.5mm}
         
    \caption{This example entry shows which information we consider for this shared task. First, we incorporate the \textit{semantic} information as the text of queries and arguments. Secondly, we use the demographic and socio-cultural properties (\textit{perspective}) of argument authors or users, including \textit{age}, \textit{gender}, or \textit{political attitude}.}
    \label{fig:example}
\end{figure}

\begin{figure}[t]
    \centering
    \includegraphics[width=0.40\textwidth]{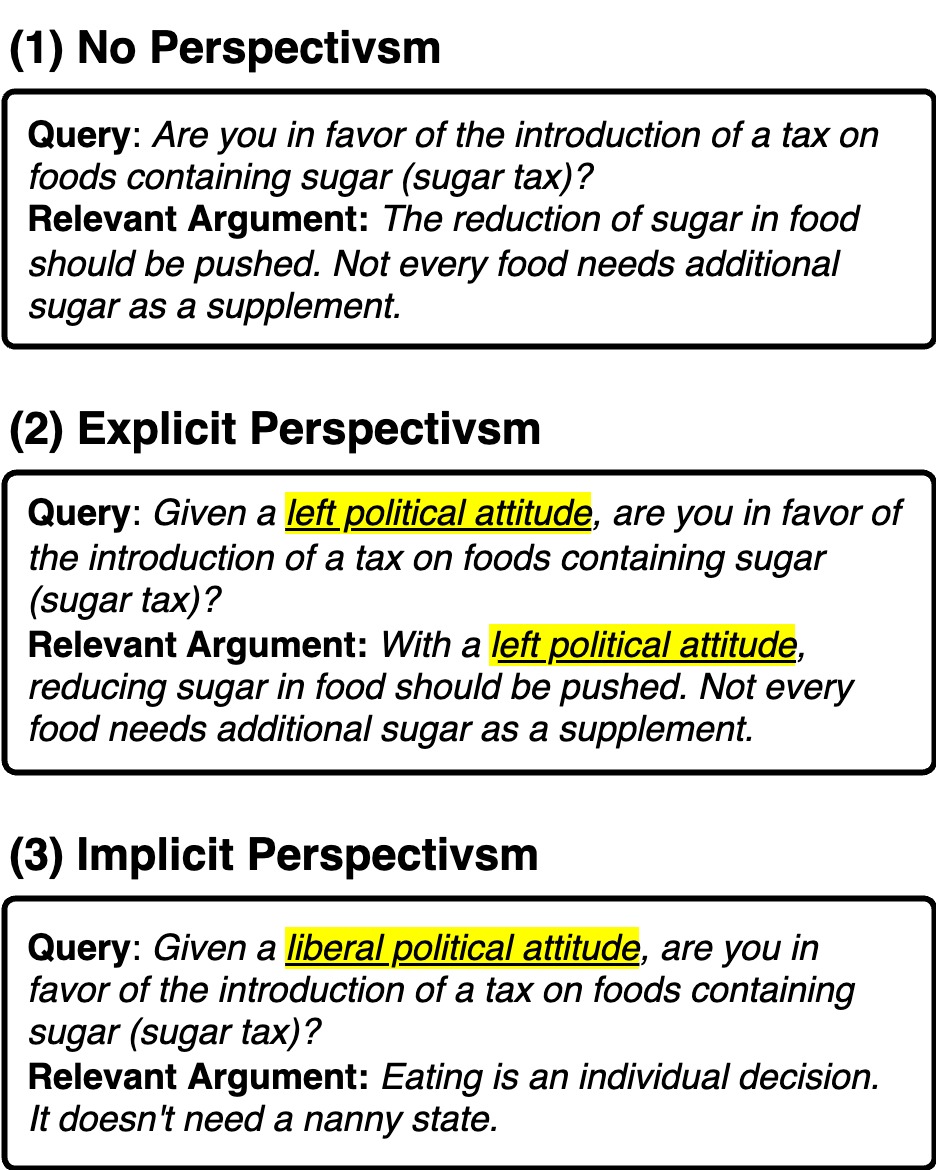}
    
         \vspace{-1.5mm}

    \caption{Examples of query and a relevant argument for the three scenarios: (1) no perspectivism without \underline{\textit{socio} variables}; (3) explicit perspectivism with \underline{\textit{socio} variable} in query and argument; (3) implicit perspectivism with \underline{\textit{socio} variable} only in the query.}
    \label{fig:scenarios}
\end{figure}

Addressing this research gap, we introduce the \textit{Shared Task on Perspective Argument Retrieval} (\autoref{sec:task}). 
Incorporating the \textit{perspective} of authors and readers (\autoref{fig:example}), we aim to foster \textbf{personalization} by retrieving arguments that match individual perspectives beyond their \textit{semantic} alignment and \textbf{reduce polarization} by enabling individuals to compare and contrast arguments from their own and other perspectives.
Therefore, we present a novel multi-lingual dataset (\autoref{sec:data}) providing demographic and socio-cultural (\textit{socio}) profiles of argument authors or readers for German, French, and Italian. 
In this context, relevant arguments are semantically aligned with a given query and match the specific \textit{socio} variables provided in the query.
We use three scenarios (\autoref{fig:scenarios}) to disentangle the effect of \textit{perspectivism}:
\begin{itemize}
    \item \textbf{No Perspectivism}: The vanilla argument retrieval scenario as a reference.
    \item \textbf{Explicit Perspectivism}: Verifying whether retrieval systems can achieve personalization regarding \textit{socio} variables when mentioned in the query and argument corpus.
    \item \textbf{Implicit Perspectivism}: Assessing the solely text-based personalization abilities of retrieval systems as \textit{socio} variables are only given in the query, and we expect systems to exploit fine-grained socio-linguistic variations between authors with different profiles.
\end{itemize}

With this shared task, we aim to examine how retrieval systems can exploit the latent influence of demographic and socio-cultural profiles, such as age or political attitude, and how they are biased regarding over- or underrepresented groups (like different age groups).
Current approaches in computational argumentation tend to prioritize majority groups and marginalize minority groups \citep{spliethover-wachsmuth-2020-argument,holtermann-etal-2022-fair}.
To fulfill these objectives, we adopt a fine-grained and comprehensive evaluation protocol and assess the performance of submitted argument retrieval systems in two tracks: \textbf{relevance} and \textbf{diversity} (\autoref{sec:evalation}).
The retrieval system should provide top-$k$ arguments that are highly relevant to the query and simultaneously diversify varying demographic and socio-cultural profiles.
Therefore, we rely on prior work enforcing diversification across stances in retrieved arguments \citep{CherumanalSSC21}.
With this shared task and results from the six participating teams, we address the following research questions:

\paragraph{RQ1: Can argument retrieval systems encode socio-cultural variables?}
Results (\autoref{sec:results}) reveal substantial challenges in encoding perspectives and successfully achieving personalization.
Systems struggle to capture fine-grained socio-linguistic features without explicit profile mentions. 
Moreover, there is a lack of suitable metrics for evaluating relevance, diversity, and fairness.

\paragraph{RQ2: Are argument retrieval systems biased regarding specific socio-cultural variables?}
While retrieval systems primarily follow the corpus bias, in-depth analysis (\autoref{sec:analysis}) shows that they balance gender bias but increase age group bias.

\paragraph{RQ3: How do argument retrieval systems generalize when switching the perceiving perspective from authors to readers?}
Perceiving perspective causes substantial performance drops (\autoref{sec:results}), as readers select arguments according to their political standing (attitude and important issue) but not regarding their demographic ones, like age or denomination, catholic or protestant (\autoref{sec:data}).

\paragraph{Contributions}
With this shared task, we establish the task of \textit{perspective argument retrieval} and present a novel dataset covering explicitly and implicitly expressed perspectives from argument authors and readers.
A comprehensive evaluation of the submitted systems underscores the challenge of this task as the system struggles to incorporate the fine-grained linguistic influence of demographic and socio-cultural variables.
Further, while these systems mostly replicate the dataset bias, they partially overcome gender bias.
These insights call for further research to incorporate perspectivism successfully and fairly, aiming for systems providing personalization.

%As socio-demographic information has not yet been available in previous datasets, these two directions could not be explored. With the proposed task and the corresponding dataset presented in this work, a first step can be taken to close this research gap.

% Andreas
\section{Perspective Argument Retrieval}\label{sec:task}
\textit{Argument retrieval} is the task of finding top-$k$ relevant arguments $y$ within a corpus $C$ for a given query $q$ \citep{bondarenko20}.
We formulate \textit{perspective argument retrieval} as an expansion of argument retrieval to perspectivism \citep{DBLP:conf/aaai/CabitzaCB23} when finding best-matching arguments.
By considering demographic and socio-cultural (\textit{socio}) variables, we account for latent aspects of argumentation beyond semantic features, such as age, occupation, or political attitude. 
Concretely, this shared task proposes three scenarios (\autoref{fig:scenarios}) to evaluate how argument retrieval systems can account for perspectivism.

% RQ1
% RQ2
% RQ3

\subsection{Scenario 1: No Perspectivsm}
First, we test a system's ability to retrieve relevant arguments $y$ solely using semantic features of arguments in the corpus $C$ and the query $q$ without any \textit{socio} variables.
This scenario represents the classical retrieval setup as reference performance.

\subsection{Scenario 2: Explicit Perspectivsm}
Second, we add \textit{socio} variables to both corpus and query to test whether a retrieval system can consider \textit{socio} variables when explicitly given, like \textit{left political attitude}.
This scenario simulates adopting the retrieval stage to specific perspectives while retaining the argument retrieval performance.
For this shared task, we only consider one \textit{socio} variable at a time to test the effect of considering them in isolation.
Consequently, this scenario is computationally heavy as systems must encode the argument corpus for every considered \textit{socio} variable in the queries. 
For example, when querying for a specific \textit{socio} properties, like the age group \textit{18-34}, the corpus must be encoded with the corresponding \textit{socio} property of the arguments, such as the specific age group.

\subsection{Scenario 3: Implicit Perspectivsm}
This third scenario is similar to the second one (\textit{explicit perspectivism}), but we only add \textit{socio} variables to the query, like \textit{liberal political attitude}. 
It is better aligned with real use cases as \textit{socio} variables of arguments are often not given and represent true \textit{personalization}. 
As a result, we aim for a retrieval system with which we can account for latent encoding of \textit{socio} variables within arguments.
Furthermore, this scenario is computationally more efficient than the \textit{explicit} one because arguments do not need to be encoded more than once.
% Andreas
\section{Data}\label{sec:data}
In the following, we outline the data used for this shared task, involving the data source (\autoref{subsec:source}), the used demographic and socio-cultural variables (\autoref{subsec:socios}), the composition of the argument corpus and the queries (\autoref{subsec:composition}).

\subsection{Source}\label{subsec:source}
We conduct this shared task with data provided by the voting recommendation platform SmartVote\footnote{\href{https://www.smartvote.ch/}{https://www.smartvote.ch/}} from the Swiss national elections of 2019 and 2023.\footnote{Data of the 2019 elections were used in previous works, like \citep{DBLP:conf/swisstext/VamvasS20} for multi-lingual stance detection.}  
This platform provides voting suggestions based on a questionnaire that politicians and voters fill out.\footnote{More information about the questionnaire and scientific methodology available \href{https://web.archive.org/web/20240726155426/https://www.smartvote.ch/en/wiki/methodology-questionnaire?locale=en_CH}{online}.}
In it, politicians can argue why they are in favor or against specific political issues.
Concretely, we use these arguments formulated by politicians, the political issue addressed by one of these arguments, the stance of an argument regarding the political issue, and the \textit{socio} variables of the politicians (authors) who formulated these arguments. 
We pre-process the data following \citep{DBLP:conf/swisstext/VamvasS20} and remove arguments with less than 50 characters, including URLs, or are not formulated in German, Italian, or French. 
After this filtering, we compose around 41k arguments written by 3.8k unique politicians regarding 266 distinct political issues in German, Italian, and French and an average of 15.7 arguments per person. 
We use these arguments to form the retrieval corpus $C$ and use the political issues as queries $q$, either with (\textit{explicit \& implicit perspectivism}) or without (\textit{no perspectivism}) corresponding \textit{socio} variables of the authors.
Given a query $q$, we define relevant arguments as those written by politicians to address the specific political issue of $q$.
Note that this is a binary assignment without any fine-grained relevance measure.

\subsection{Demographic and Socio-Cultural Variables}\label{subsec:socios}
\begin{figure*}[t]
    \centering
    \includegraphics[width=0.95\textwidth]{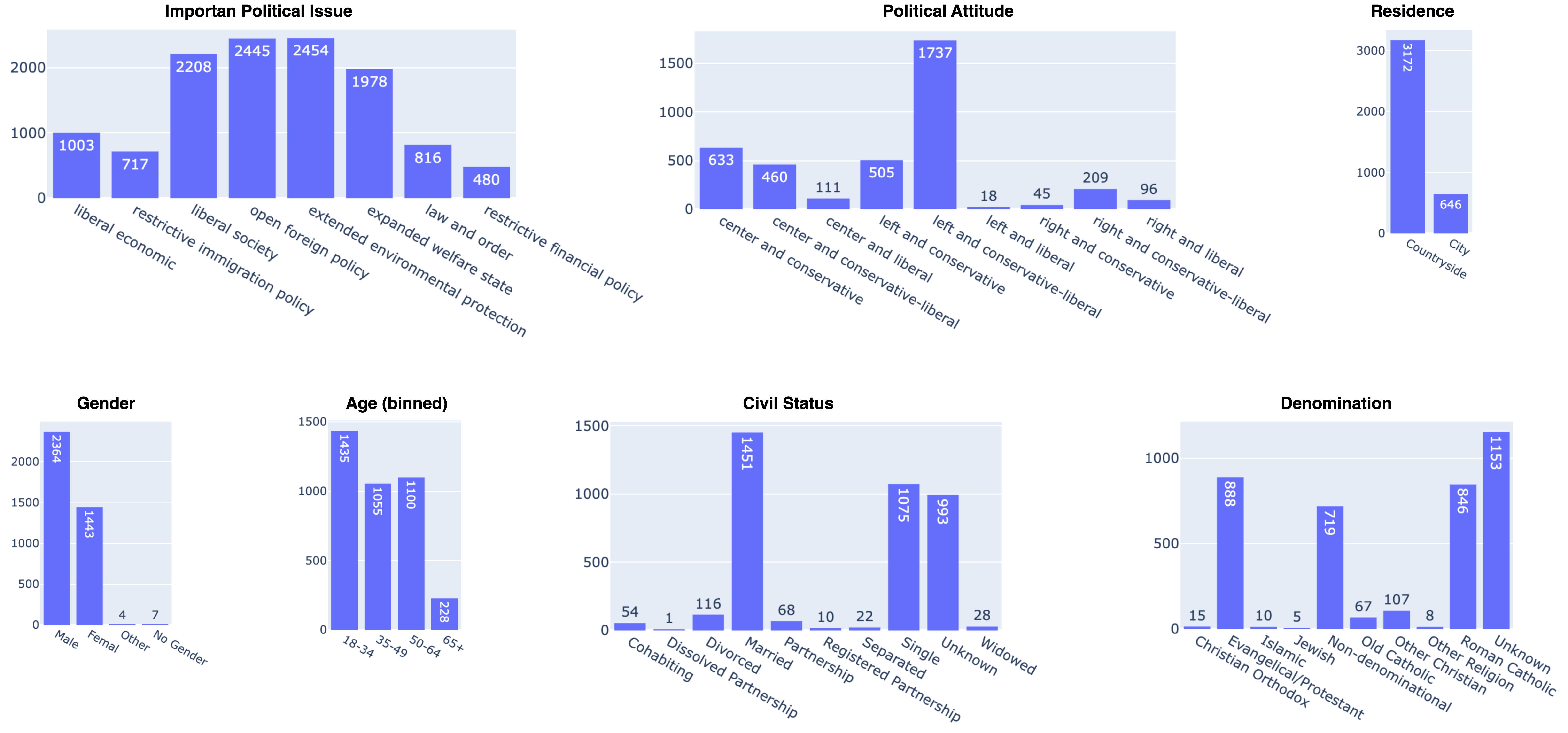}

         \vspace{-2.5mm}

    \caption{Distribution of the politicians' different demographic and socio-cultural variables: important political issues, political attitude, residence, gender, age (binned), civil status, and denomination.
    Note, that one person can have more than one important political issue.}
    \label{fig:socios}
\end{figure*}

We use \textit{socio} variables of the politicians (authors) who formulated the arguments.
\autoref{fig:socios} provides an overview of them, including the following personal information: \textit{gender}, \textit{age (binned)}, \textit{residence} (either city or countryside), \textit{civil status}, and \textit{denomination}. 
Further, SmartVote provides a \textit{SmartMap} ranking of the politicians on a left/middle/right and conservative/liberal dimension based on answers to the full questionnaire.\footnote{More information about the SmartMap available \href{https://web.archive.org/web/20240726155426/https://www.smartvote.ch/en/wiki/methodology-questionnaire?locale=en_CH}{online}.}   
We combine (binning) these two dimensions into a single variable \textit{political attitude}.
Finally, SmartVote indicates, with a \textit{SmartSpider}, the \textit{important political issues} of a person based on the answered questionnaire.\footnote{More information about the SmartSpider available \href{https://web.archive.org/web/20240726155435/https://www.smartvote.ch/de/wiki/methodology-smartspider}{online}.}
One person can have more than one out of eight \textit{important political issues}: open foreign policy, liberal economic, restrictive financial policy, law and order, restrictive immigration policy, extended environmental protection, expanded welfare state, and liberal society.
The insights of \autoref{fig:socios} show the demographic bias of politicians, such as living on the countryside, identifying as male, or being married.

\begin{figure}[t]
    \centering
    \includegraphics[width=0.4\textwidth]{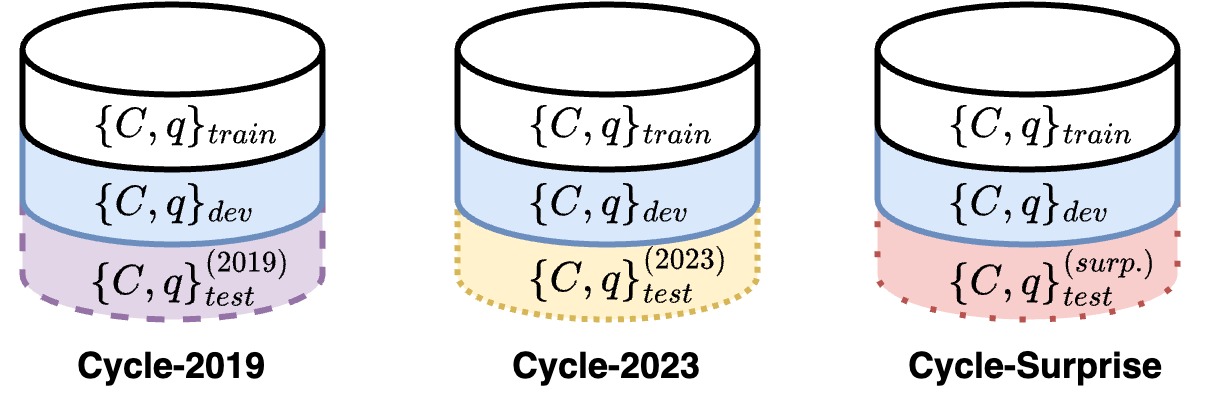}

         \vspace{-2.5mm}

    \caption{Overview of train, dev, and test argument corpora ($C$) and queries $q$ for the three evaluation cycles dataset (\textit{2019}, \textit{2023}, \textit{surprise})}
    \label{fig:cycles}
\end{figure}

\subsection{Dataset Composition}\label{subsec:composition}
We compose three versions of the dataset with distinct test sets to run three different evaluation cycles (\autoref{fig:cycles}) covering (1) data from the 2019 election, (2) data from the 2023 election, and (3) surprise data.
For every cycle, a dataset consists of distinct train, dev, and test queries ($q_{train}$, $q_{dev}$, and $q_{test}$) along with a corpus of arguments, $C=\{C_{train}, C_{dev}, C_{test}\}$.
We include all relevant arguments for at least one query within the corresponding part of the corpus. 
While train $q_{train}$ and dev queries $q_{dev}$ remain the same, we use distinct test queries ($q_{test}^{(2019)}$, $q_{test}^{(2023)}$, and $q_{test}^{(surp.)}$) for every cycle.
Subsequently, the arguments ($C_{train}$, $C_{dev}$) remain the same, but the test part of the corpus ($C_{test-2019}$, $C_{test-2023}$, and $C_{test-surp.}$) is updated with the specific arguments which are relevant for the corresponding test queries. 
Note that for a given $q_i$ we expect to retrieve arguments from the full corpus $C$.
Since every query has a German, French, and Italian version, we include a separate one for each language.
However, we consider arguments for any language as relevant.
For example, the German and French versions of $q_i$ share their relevant arguments $y$.

\paragraph{Train and Dev} 
We use 35 and 10 distinct political issues from the 2019 election as train and dev queries ($q_{train}, q_{dev}$) and include ~21k arguments and ~2k ones for dev in the corpus ($C_{train}$, $C_{dev}$).

\paragraph{Test Cycle-2019}
During the first evaluation cycle, we use an additional 15 distinct political issues from the 2019 election as test queries ($q_{test}^{(2019)}$).
The corresponding corpus ($C_{test}^{(2019)}$) consists of ~6k arguments. 
With this test set, we evaluate the retrieval performance given the topic shift between train, dev, and test queries/arguments as they cover distinct political issues. 

\paragraph{Test Cycle-2023}
For the second evaluation cycle, we select 62 distinct political issues from the 2023 election for testing ($q_{test}^{(2023)}$) along with ~13k arguments ($C_{test}^{(2023)}$).
This second cycle saturates the topic shift between train, dev, and test sets as new topics gained political relevance between 2019 and 2023, like Corona or the war in Ukraine. 

\paragraph{Test Cycle-Surprise}
Finally, we conduct an annotation study to assess whether retrieval systems generalize when we change the perceiving perspective from authors to readers (\textbf{RQ3}). 
Concretely, this study covers 27 political issues and 20 arguments from the 2023 election answering these issues.
We conducted this annotation study with 22 crowd workers recruited from prolific.
More details about their selection, background, and payment are in Appendix \autoref{app:annotation_study}. 
During annotation, we present the annotators with 20 arguments for every political issue and ask them to select those they intuitively perceive as relevant for answering the presented issue.
Along with this selection, we collect the \textit{socio} profile of the annotators as done by SmartVote for the authors. 
Based on these annotations, the test portion of the argument corpus ($D_{test}^{(sure.)}$) for this cycle consists of 540 arguments (20 arguments for every 27 political issues).
Further, we use the 27 political issues and the \textit{socio} profiles of the annotators to form the test queries ($q_{test}^{(sure.)}$).
Noteworthy, we find that annotators perceive arguments as relevant when they share the \textit{political spectrum} and \textit{important political issues} with the authors of the arguments (see \autoref{fig:personalization} in Appendix \autoref{app:annotation_study}).
\section{Evaluation}\label{sec:evalation}
We employ a two-folded evaluation to comprehensively measure the retrieval quality for all three scenarios. 
Concretely, we distinguish between \textbf{relevance} and \textbf{diversity}.

\paragraph{Relevance}
With relevance, we focus on the ability of a retrieval system to select relevant candidates given the query, for example, all arguments addressing the queried issue for the baseline scenario or arguments that additionally match specific socio-cultural properties for explicit or implicit perspectivism. 
Following previous work \cite{bondarenko20,bondarenko22,thakur}, we use the Normalized Discounted Cumulative Gain (nDCG@) and precision@ metric. 
Compared to precision, nDCG has the advantage of taking the position of the ranked items into account.
Therefore, the metric places greater emphasis on higher-ranked arguments.

\paragraph{Diversity}
Using diversity, we account for the influence of perspectivism in the evaluation by measuring whether a retrieval system proposes balanced arguments regarding the stance distribution and the authors' diverse socio-cultural properties (such as gender or political attitude). Following previous studies regarding fairness in argument retrieval systems \cite{CherumanalSSC21}, we calculate alpha-nDCG@ for each available property separately and average them afterwards. This metric accounts for relevance and diversity by assessing whether an item is relevant and introduces a new perspective compared to the previous one. Consider the following example: a system retrieved a list of arguments relevant to a given issue, and we aim to evaluate diversity for political attitude. The metric would prefer the arguments to be sorted like this ['liberal', 'conservative', 'left', 'conservative'] over ['conservative', 'conservative', 'liberal', 'left']. An optimal ranking ensures that all relevant perspectives for a corresponding socio-cultural variable are represented among the top-ranked arguments.
Note that these conditioned properties are withheld when evaluating diversity since we condition specific socio-demographic properties in the query for explicit or implicit perspectivism (scenarios 2 and 3). 

As a second metric, we look at the Normalized discounted KL-divergence, introduced as a metric of \textit{unfairness} \cite{CherumanalSSC21}. This metric evaluates the fairness of the ranking by comparing the distribution of a protected group (e.g. what is the proportion of arguments by female authors when looking at the property 'gender') in the top-k ranked items against a gold standard proportion (what is the proportion of arguments by female authors in the whole corpus?). The divergence is calculated at specified cut-off points and then averaged, with each point discounted by the logarithm of its rank, to assess how well the ranking
reflects the representation of the protected group. In this case, the relevance of an argument is not considered; instead, the metric can reveal biases against specific groups. For example, it can show whether systems disproportionately favor dominant groups in the top arguments.

\paragraph{Final Ranking}
We evaluate the performance at four different values of k [4, 8, 16, 20] and calculate the average performance across these k values. This evaluation is conducted for the three scenarios across three different test sets, resulting in 9 scores for relevance and 9 for diversity. The final rank is determined by averaging these nine scores.

% Neele
\section{Submissions}\label{sec:submissions}
In the following, we summarize the baseline (\autoref{subsec:baseline}) and the submitted systems (\autoref{subsec:submissions}).
Further, we elaborate on the unique ideas incorporated by the participants.

\subsection{Baseline Systems}\label{subsec:baseline}
We provide two baseline systems (\texttt{BM-25} and \texttt{SBERT}) to evaluate how simple retrieval systems perform without being optimized for any perspectivism.

\paragraph{Baseline \texttt{BM-25}:} the BM-25 ranking algorithm ranks arguments based on lexical overlap. It is computed using tf-IDF but also accounts for document length \cite{bm25}.

\paragraph{Baseline \texttt{SBERT}:} we use SBERT \cite{sbert} and the model \texttt{paraphrase-multilingual-mpnet-base-v2} to encode the query and the arguments from the corpus, ranking them based on cosine similarity. We encode the socio-cultural variables within the query for the perspectivist approaches. In Scenario 2, we concatenate the entire socio-cultural profile with each argument in the corpus.

\subsection{Submitted Systems}\label{subsec:submissions}
This shared task received submitted systems from six teams: \texttt{twente-bms-nlp} \citep{twente}, \texttt{sövereign} \citep{soevereign}, \texttt{GESIS-DSM} \citep{gesis}, \texttt{turiya} \citep{turiya}, \texttt{xfact} \citep{xfact}, and \texttt{boulderNLP} (no system paper submitted).
Some systems did not submit results for all three scenarios but instead focused on one or two (e.g., no perspectivism and explicit). 
We summarize and elaborate on the specific techniques of these systems, including embedding strategies, candidate filtering \& re-ranking, using LLMs, or using auxiliary classification tasks.

\paragraph{Embedding queries and arguments}
All systems used SBERT \cite{sbert} to encode queries and arguments and retrieve an initial set of relevant arguments by calculating the cosine similarity between query and corpus embeddings. 
Additionally, \texttt{twente-bms-nlp} uses cross-encoding LMs to re-rank the top 50 arguments, and \texttt{turiya} fuses rankings obtained with mono- and multi-lingual embeddings, once using KNN and once cosine similiarty for ranking.
Only \texttt{xfact} further fine-tunes SBERT to align the semantic representations of relevant arguments and corresponding queries. 
They use other arguments as negative examples and optimize the representations with multiple negative ranking losses.

\paragraph{Filtering out irrelevant arguments}
Most teams filter relevant candidates before (re)ranking: for scenario 2, they hard-filter arguments that explicitly match the socio-demographic variable in the query.
\texttt{twente-bms-nlp} filters arguments that appear relevant in the training set to reduce the candidate pool arguments that likely match the political issue in test queries, as there is no overlap between train and test queries.
\texttt{xfact} filter arguments that had no overlap between keywords of the query and keywords of the arguments. 

\paragraph{Re-ranking top k arguments}
Some teams retrieve a larger list of relevant candidates and then adopt complex strategies to re-rank the top-k arguments due to their high weight in the evaluation. 
These strategies often include training a specific classifier, e.g., \texttt{turiya} trained two classifiers, one binary to assign a relevance label (0 or 1) given query and argument, and one to select the more relevant argument out of two.
\texttt{sövereign} prompt an LLM to generate relevance scores given query and a list of the top 50 retrieved arguments. 
For perspectivism, they include instructions to determine whether the given socio-demographic property matches the arguments.

\paragraph{Additional use of LLMs}
Four out of five teams use LLMs at some point in their pipeline. 
Two teams (\texttt{xfact} and \texttt{GESIS-DSM}) explore the idea of 'prototypes' or 'anchors' and automatically generated arguments given a specific query. 
\texttt{GESIS-DSM} uses the generated arguments as a reference anchor to re-rank the relevant candidates with SBERT.
For perspectivism, this generated argument should represent specific demographic properties.
\texttt{xfact} utilizes LLM to generate prototypical and diverse arguments in response to a query. 
These generated arguments serve as centroids in a clustering process designed to identify relevant arguments within the corpus.
The approach ensures that the retrieved arguments are relevant and exhibit greater diversity by creating a variety of arguments.

\paragraph{Additional classification tasks to identify relevant arguments}
Several teams train additional classifiers to enhance system performance, whether to improve the identification of relevant arguments or retrieve arguments matching specific socio-cultural variables.
\texttt{xfact} uses stance detection as an auxiliary task to improve the system's ability to detect whether an argument matches a query.
In the final stage, the classifier's confidence level is used as a cutoff radius to selectively refine the set of relevant arguments when comparing their distance to the centroids generated by the LLM.
\texttt{sövereign} uses a logistic classifier to learn a more informed relevance score for re-ranking: the classifier incorporates cosine similarity, a demography matching score, and a topic frequency score as features.
\texttt{twente-bms-nlp} and \texttt{GESIS-DSM} investigate whether classifiers can learn to predict the values for certain socio-cultural variables from the arguments. 
Both compared the performance of classifiers using semantic content against linguistic (style) features. 
\texttt{twente-bms-nlp} find that the classification of the different values is challenging but can improve the final results of the system using a classifier that predicts whether an issue is important for an author based on a semantic representation of the argument. 
\texttt{GESIS-DSM} finds that semantics were less predictive of differences between groups of different socio-cultural variables and instead can retrieve a better re-ranking when using several linguistic style features as predictors.

\section{Results}\label{sec:results}
In the following section, we discuss the results of the submitted systems focusing on \textbf{RQ1}. 
Additional detailed discussions regarding single scenario, evaluation cycle, and top-$k$ are in Appendix \autoref{subsec:detailed-results} and \autoref{app:subsec_top-k}.

\paragraph{Relevance and diversity agree but not with fairness.}

\autoref{leaderboards} shows each track's final leaderboards. Both tracks (relevance and diversity) share the same team rankings.
All teams outperform the \texttt{SBERT} baseline when they submitted for all scenarios (\texttt{xfact} and \texttt{boulderNLP} have only submitted 6 / 3 prediction files, leading to lower scores.)

\begin{figure}[t]
    \centering
    \includegraphics[width=0.48\textwidth]{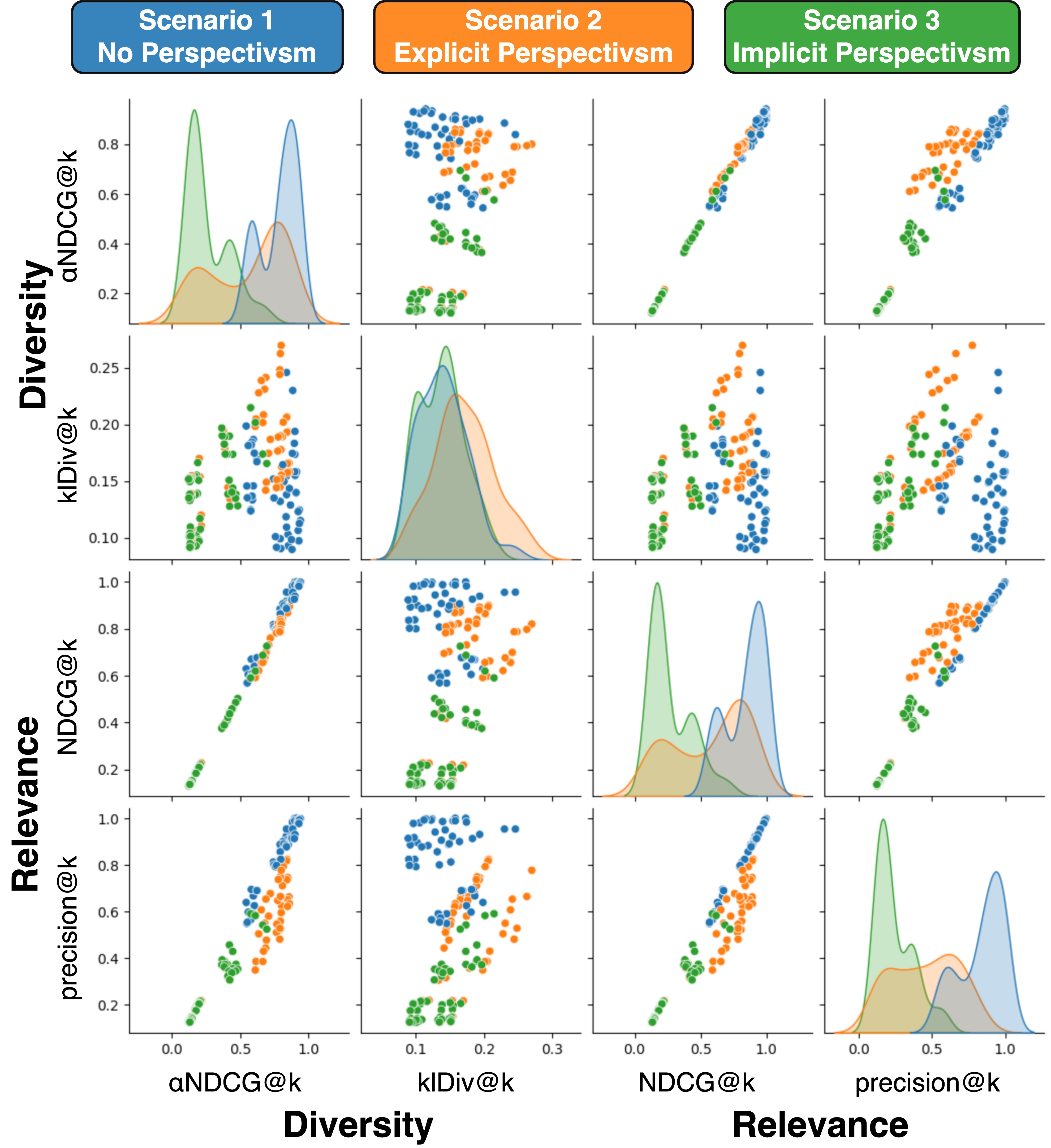}

     \vspace{-2.0mm}
     
    \caption{Performance overview regarding the four measured metrics and their relation. The color indicates the specific scenario.}
    \label{fig:metrics}
\end{figure}

Next, we compare the metrics representing relevance ($\text{NDCG}@k$), diversity ($\alpha\text{NDCG}@k$), and fairness ($\text{klDiv}@k$). 
Relevance and diversity are highly correlated, but diversity scores are lower than relevance, showing that no system perfectly diversifies its top-k arguments.
Compared with fairness ($\text{klDiv}@k$), relevance and diversity are weakly correlated $\rho=0.13$.
Ideally, we expect a correlation of $\rho=-1$ as $\text{klDiv}@k=0$ would represent a perfectly fair system and $\text{NDCG}@k=1$ and $\alpha\text{NDCG}@k=1$ indicates perfect relevance and diversity. 
\autoref{fig:metrics} confirms these patterns in more detail with results across every $k$, evaluation cycle, scenario, and team.
Furthermorea , these insights are consistent with \citet{CherumanalSSC21}, which states that these metrics are not equivalent and measure different dimensions.

\begin{comment}
\paragraph{Relation of measured metrics}
We further analyze in \autoref{fig:metrics} how the different measures relate to each other based on the results for every $k$, evaluation cycle, scenario, and team.
NDCG@k, precision@k, and $\alpha$NDCG@k are highly and significantly correlated (Pearson $\rho>0.95$).
In contrast, klDiv@k marginally correlates with the other metrics ($rho$ of 0.20 to 0.30) as it does not consider the relevance of the candidate arguments.
These patterns are consistent across scenarios. 
Exceptionally, precision@k shows a weaker correlation for the second scenario (\textit{explicit perspectivism}).
With these results in mind, we focus on $\alpha$NDCG@k in the subsequent discussion as it balances relevance and diversity.  
\end{comment}

\begin{table}[]
\small
\resizebox{0.48\textwidth}{!}{%
\begin{tabular}{@{}lcccc@{}}
\toprule
       & \multicolumn{2}{c}{\textbf{Relevance}} & \textbf{Diversity} & \textbf{Fairness} \\ \midrule
                         & \textbf{ndcg@k}      & precision@k     & \textbf{ \(\alpha\text{NDCG}@k\)}       & klDiv@k  \\\midrule
\textbf{twente-bms-nlp}  & 70.7                & 63.4           & 67.2                  & 16.7         \\
\textbf{sövereign}       & 63.2                & 56.1           & 60.1                  & 15.9         \\
\textbf{GESIS-DSM}       & 60.7                & 54.3           & 57.9                  & 15.0         \\
\textbf{turiya}         & 51.8                &       -     & 49.5                 &    -     \\
\textbf{sbert} & 44.5                & 42.7           & 41.9                  & 0.136         \\
\textbf{xfact}         & 41.7                & 40.0           & 39.4                  & 0.136         \\
\textbf{boulderNLP}      & 29.2                &       -     & 27.1                  &   -      \\
\textbf{bm25}  & 19.5                &       -     & 18.5                  &   -     \\ \bottomrule
\end{tabular}
}
\caption{Final result of the shared task regarding relevance (NDCG and precision), diversity ($\alpha\text{NDCG}$), and fairness (klDiv).
}
\label{leaderboards}
\end{table}

% - which system is best in which test set? difference between diversity and relevance? Different rankings for different k-s?

% how difficult is no perspective vs explicit vs implicit? 

% how good is the baseline?

\begin{figure}[t]
    \centering
    \includegraphics[width=0.40\textwidth]{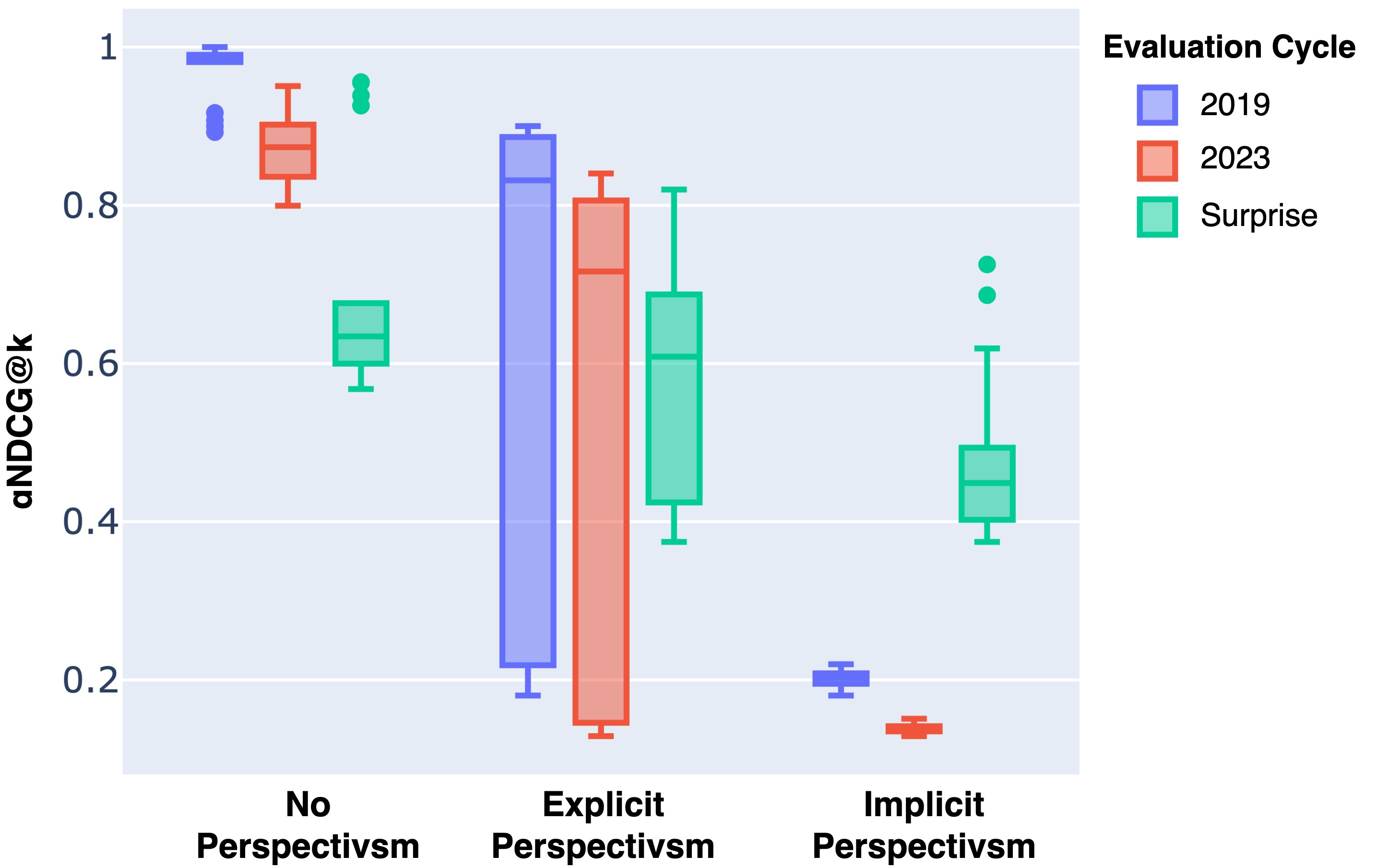}

    \vspace{-2.0mm}

    \caption{Performance comparison of the three evaluation cycles (color) regarding the three scenarios (x-axis) for diversity (y-axis, $\alpha$NDCG@k).}
    \label{fig:box}
\end{figure}

\paragraph{Considering Perspectivism is difficult.}
We analyze the performance differences between the three scenarios.
\autoref{fig:metrics} and \autoref{fig:box} reveal that considering no \textit{socio} variable (scenario one) performs the best across all test sets of the three evaluation cycles.
Comparing scenarios one with two and three (considering perspectivism \textit{explicit} or \textit{implicit}) highlights the challenges of incorporating \textit{socio} variables in the retrieval stage.
This becomes even more apparent when comparing scenarios two and three.
While considering \textit{socio} variables in the query and corpus (scenario two) results in higher performance, it crucially requires more computing.
In contrast, the more efficient approach of considering \textit{socio} variable only in the query (scenario three) causes significant performance degradation.
Thus, existing retrieval systems show crucial limitations in taking into account \textit{perspectivism}, either \textit{explicit} or \textit{implicit}.
Particularly, they need the signal of the \textit{socio} variable within the query and corpus.
Further analysis of the participating teams reveals that the implicit difference between arguments of distinct \textit{socio} variables is more stylistic than semantic.
\textbf{As a result, we see the need to build retrieval systems accounting for such fine-grained socio-linguistic variations to consider perspectivism accurately and efficiently.}

\paragraph{Temporal shift reduces retrieval performance.}
We analyze the temporal effect when comparing results from the test sets covering the 2019 (blue) and 2023 (red) elections. 
\autoref{fig:box} shows that this temporal shift has a crucial effect on the retrieval performance for all three scenarios. 
We see this shift mainly as semantic as we consider political issues regarding freshly raised topics like Corona or the war in Ukraine.

\paragraph{The importance of the perceiving perspective.}
With the third evaluation cycle, we focus on \textbf{RQ3} and analyze how the retrieval system handles queries when the receiving perspective of the arguments changes. 
We see systems struggling when comparing the authors' (2019 and 2023) with the voters' perspective (surprise), in particular for the first and second scenarios. 
While these results provide first insights, more extensive studies are required to cover the same demographic variance as in the 2019 and 2023 test sets. 
%However, more detailed results suggest that the retrieval systems provide more improvement over the \texttt{SBERT} baseline for \textit{socio} variables where voters' and authors' profiles match. 
Further, this smaller corpus is also reflected in the better performance of the third test set on the third scenario (\textit{implicit perspectivism}).

% Neele
\section{Analysis}\label{sec:analysis}
In the following, we focus on \textbf{RQ2} and examine whether retrieved argument candidates are biased regarding socio-cultural groups and if submitted systems compensate for such biases.
We focus on \textit{age} and \textit{gender}, known for which recent work found substantial bias in argumentation. 
Specifically, \citet{spliethover-wachsmuth-2020-argument} show that common argumentation sources (e.g. debating corpora) exhibit substantial bias regarding young ages and European-American males. 
Further, \citet{holtermann-etal-2022-fair} shows that fine-tuning LMs on argumentative data increases stereotypical bias, even if LMs exhibited a counter-stereotypical bias before tuning.
As shown in Figure \ref{fig:socios}, our dataset is biased towards specific groups, such as male and/or young authors.
We establish a random baseline by randomly sampling 20 topic-relevant arguments for every query of the implicit scenario across 10 different seeds and average the number of arguments retrieved for each group. 
Similarly, we average the performance metric.

\begin{figure}[htbp]
    \centering
    \begin{minipage}[b]{0.45\textwidth}
        \centering
        \includegraphics[width=\textwidth]{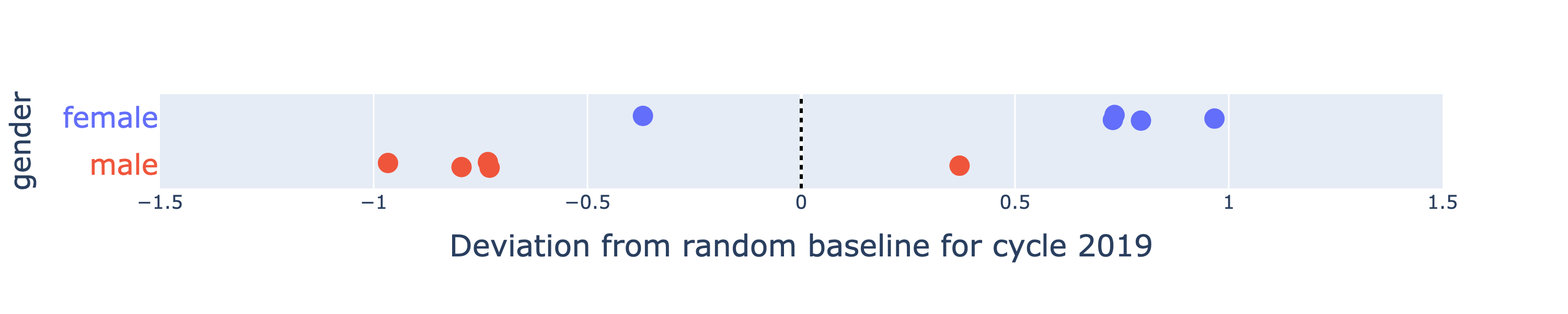}
    \end{minipage}
    
      \vspace{-2mm}
    \begin{minipage}[b]{0.45\textwidth}
        \centering
        \includegraphics[width=\textwidth]{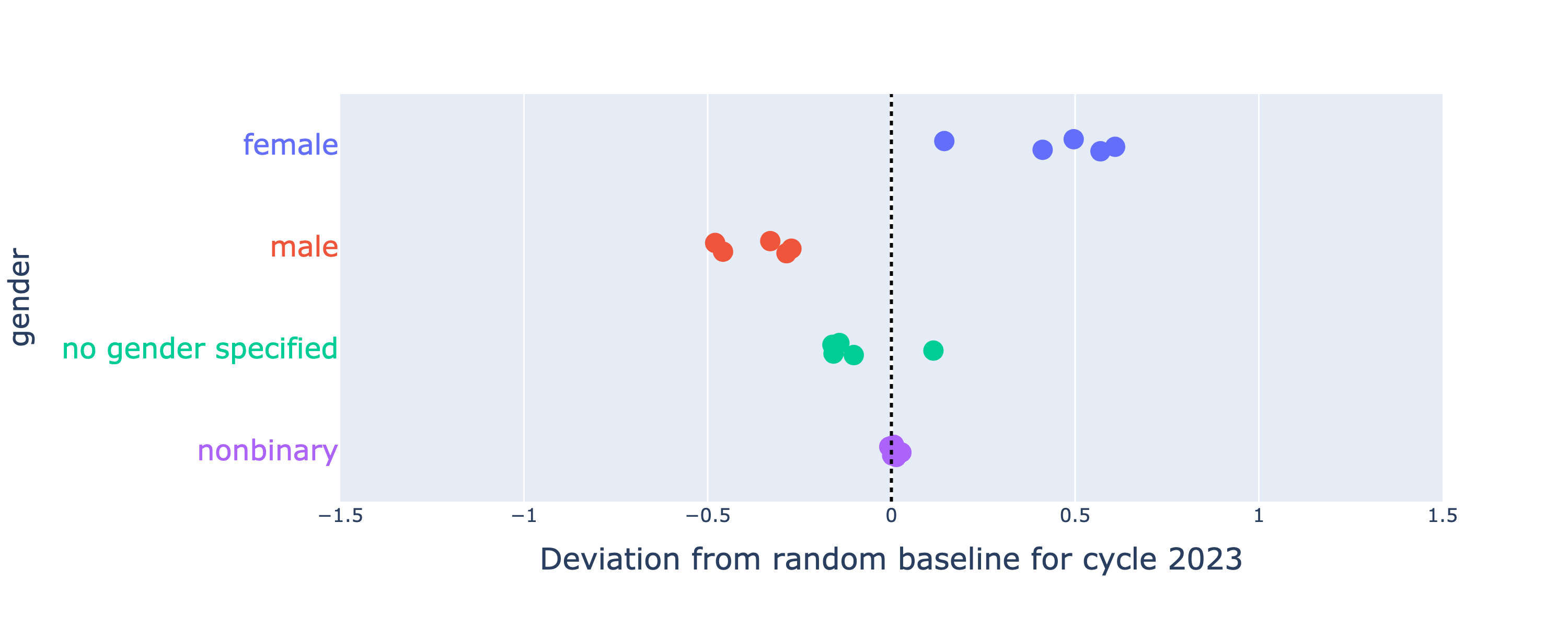}
    \end{minipage}

     \vspace{-4.5mm}
\caption{Extent of system deviation from random sampling representing each gender among the 20 most relevant arguments.}   
\label{fig:genderbias}
\end{figure}

\paragraph{Systems are biased regarding majority groups.}

We examine the 20 most relevant arguments, count how many represent the distinct group, and compute the standard deviation for each system towards the random baseline.
A negative deviation indicates that the system further reduces the representation of that group, meaning the group is less represented in the top arguments compared to its underlying distribution in the corpus. 
Conversely, a positive deviation indicates increased representation. 
In the case of a majority group, the system amplifies the bias. 

Figure \ref{fig:genderbias} shows the shift in representation for \textit{gender}, comparing the 2019 and 2023 test sets. 
We observe most systems (including the \texttt{SBERT} baseline) reducing the male bias. 
However, the top retrieved arguments still overrepresent male authors by a large margin, as the deviation is not more than one argument. Interestingly, one team reinforced the male bias in the 2019 dataset with a slight positive deviation. 
However, that system slightly outperformed the other teams in increasing the representation of other gender categories in the 2023 dataset (positive value for \textit{no gender specified}). 

\begin{figure}[htbp]
    \centering
    \begin{minipage}[b]{0.45\textwidth}
        \centering
        \includegraphics[width=\textwidth]{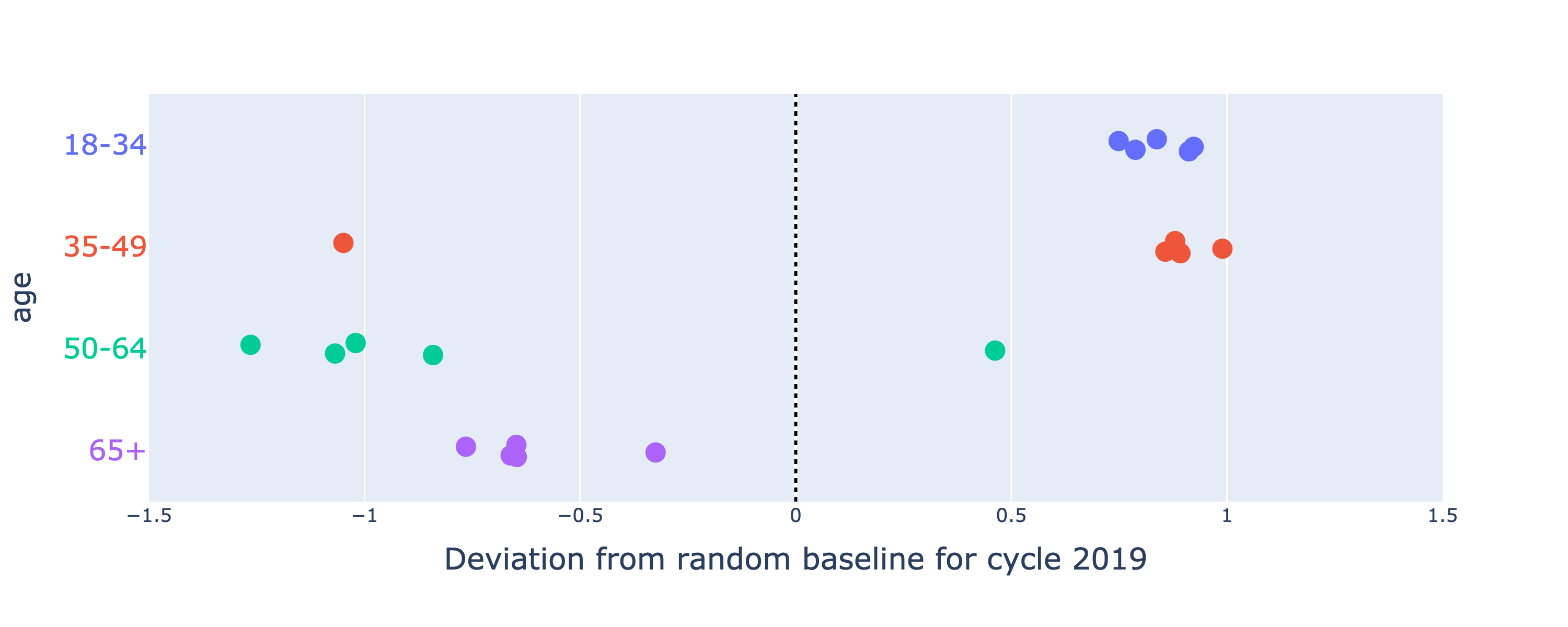}
        \label{fig:figure1}
    \end{minipage}
    
    \vspace{-7mm}
    \begin{minipage}[b]{0.45\textwidth}
        \centering
        \includegraphics[width=\textwidth]{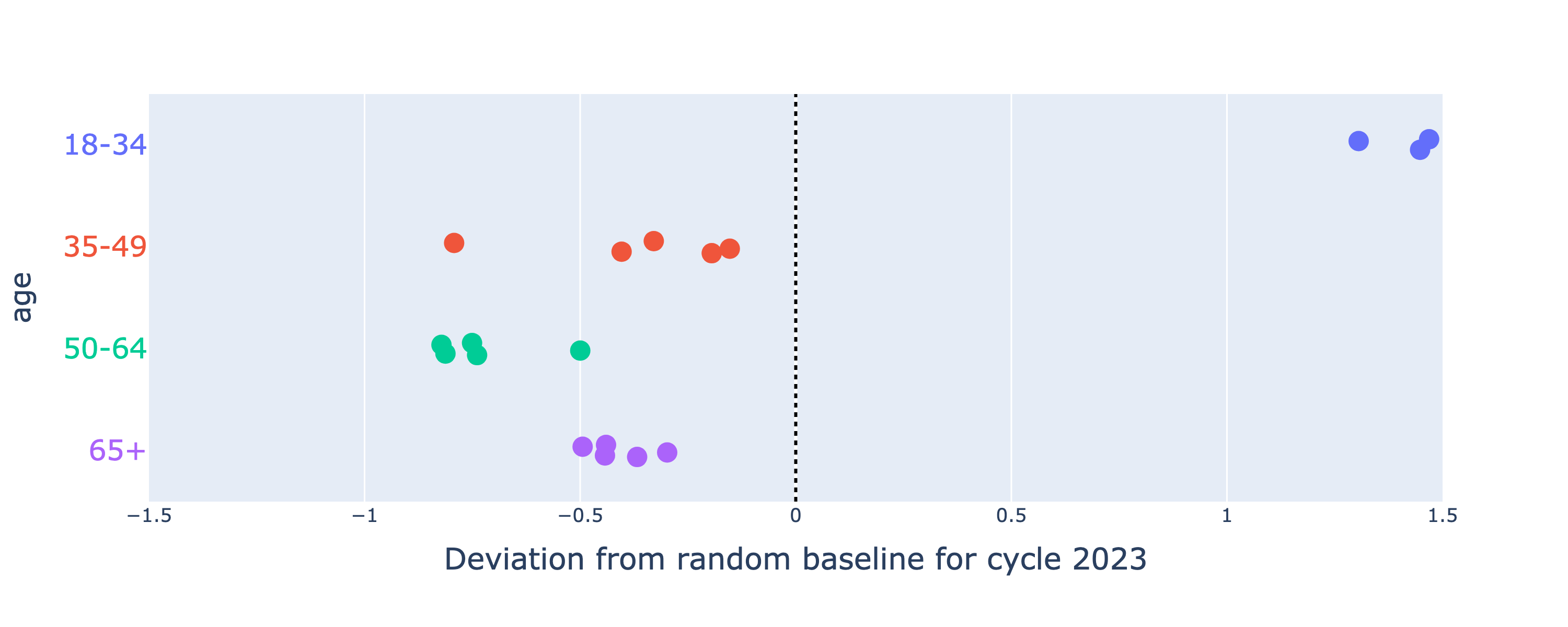}
        \label{fig:figure2}
    \end{minipage}
    
         \vspace{-6mm}

    \caption{Extent of system deviation from random sampling representing each age group among the 20 most relevant arguments.}
    \label{fig:agebias}
\end{figure}

Figure \ref{fig:agebias} focuses on different \textit{age} groups and shows that all systems reinforce a bias regarding young ages.
This is particularly true for the 2023 dataset, where systems systematically retrieve fewer arguments written by older age groups than randomly sample arguments. 
This supports general findings in NLP that older age groups are underrepresented in data and models.
Comparing the two middle-aged groups reveals that \textit{35-49} is better represented than \textit{50-64} for 2019. 
Since both age groups occur approximately equally frequently in the corpus, this indicates a stronger age bias, with the older group being significantly less well-represented.
While these findings suggest that systems are biased toward representing the majority group, they mitigate this bias more effectively for the female gender category.

%An exception is one outlier system, which represents this older group more frequently than the baseline but is the only system that represents the younger of the two age groups less frequently.

\begin{figure}[htbp]
    \centering
    \begin{minipage}[b]{0.45\textwidth}
        \centering
        \includegraphics[width=\textwidth]{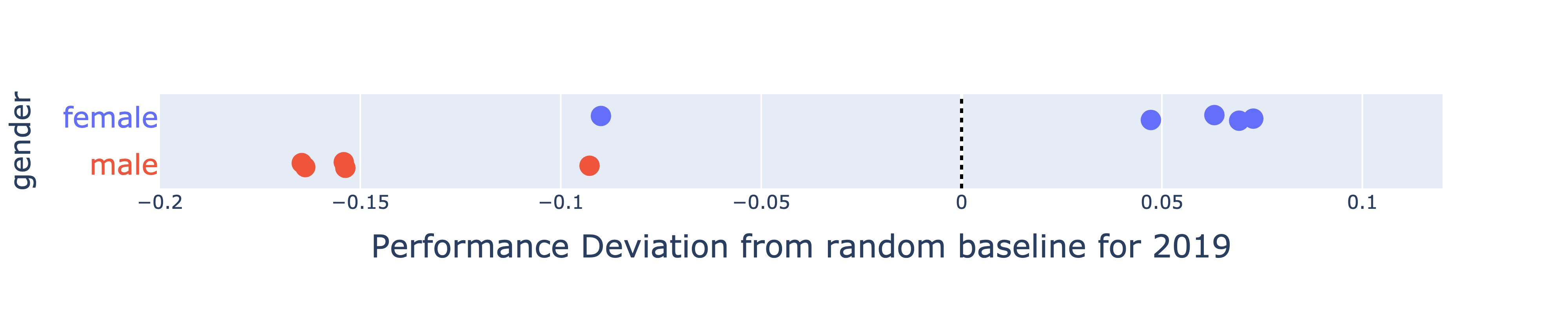}
        \label{fig:figure1}
    \end{minipage}
    
    \vspace{-7mm}
    \begin{minipage}[b]{0.45\textwidth}
        \centering
        \includegraphics[width=\textwidth]{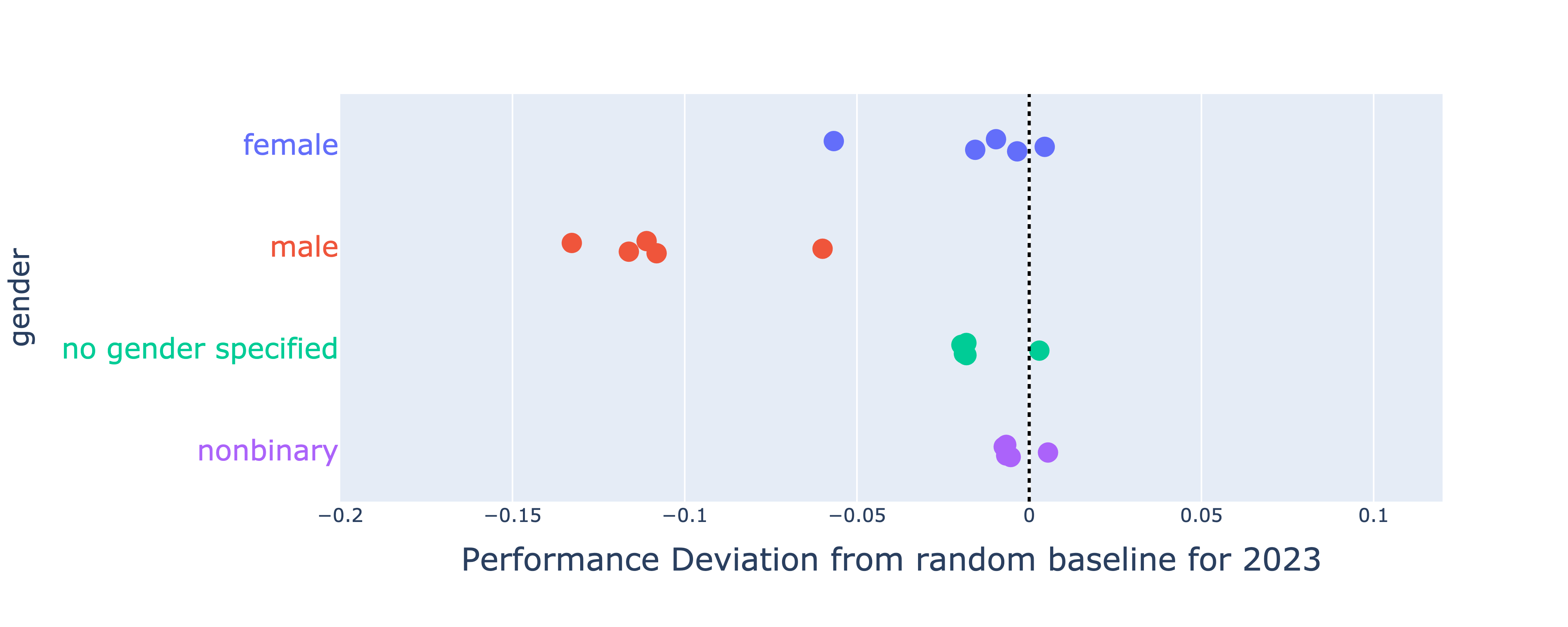}
        \label{fig:figure2}
    \end{minipage}

       \vspace{-6mm}

    \caption{Extent of system deviation from random sampling in performance from the nDCG score for different gender categories.}
    \label{fig:genderperformance}
\end{figure}

\textbf{Systems partly mitigate gender but not age bias.}
We compute each group's deviation from the system performance to the random baseline performance.
If there is no bias, the deviation for a system should be the same for each group. 
For \textit{gender}, \autoref{fig:genderperformance} shows all systems reduce the bias regarding the majority group (male gender category). For nonbinary and unspecified gender, the performance pattern is similar to representation: one system shows slight bias improvement, while the others are slightly more biased than the baseline.
The female group's performance improved for the 2019 dataset compared to the random baseline but not for the 2023 dataset.
We assume that the SBERT model has potentially seen more topics from the 2019 election and detected sub-issue-specific differences within known topics.
For example, the model could have identified specific frames used more frequently by males than females.
For \textit{age}, systems seem to agree more with the dataset distribution: younger age groups have fewer declines or even improvements (in 2023) compared to older age groups (\autoref{fig:ageperformance} in Appendix). 
Again, systems perform the worst on the \textit{50-64} age group. 

\section{Conclusion}\label{sec:conclusion}
With the \textit{Shared Task on Perspective Argument Retrieval}, we explore for the first time how argument retrieval systems align socio-cultural properties beyond topic relevance. 
Analyzing the submissions shows that semantic content alone does not distinguish between different socio-cultural groups adequately.
Instead, incorporating additional classification tasks or features is crucial for accurately matching arguments to socio-cultural characteristics. 
The subsequent analysis shows that systems overrepresent arguments from majority groups.
However, they partially mitigate these biases, such as gender bias.
By publishing data reflecting authors' and readers' perspectives, this shared task represents an initial step towards advancing argument retrieval regarding perspectivism.
This facilitates the investigation of personalization and polarization and addresses social bias and fairness in computational argumentation.

\section*{Acknowledgements}
We thank SmartVote and Summetix for their support of this shared task. 
Neele Falk is supported by Bundesministerium für Bildung und Forschung (BMBF) through the project E-DELIB (Powering up e-deliberation: towards AI-supported moderation).
Andreas Waldis has been funded by the Hasler Foundation Grant No. 21024.

\section*{Limitations}
\paragraph{Geographical Limitation}
The underlying dataset of this shared task is solely originating from Switzerland. 
While it includes distinct values of Swiss society (multilingual and through political discourse), it is limited to political issues discussed in Switzerland. 
Furthermore, the distribution of demographic and socio-cultural variables is biased regarding the Swiss population. 
For example, one expects a person in Switzerland and the United States to have a different mindset while being labeled as \textit{liberal and left}.

\paragraph{Societal Bias}
As with any usage of the language model, this work is affected by fundamental stereotypical bias injected by pre-training on past data. 
Even with a special focus during the analysis, this fact is one limitation that should be considered in any application. 

\paragraph{Appropriate Evaluation}
In a perspective-aware retrieval system, multiple metrics are essential to evaluate the system from various aspects. The diversity metric, for instance, measures whether the top arguments cover the different values of a particular \textit{socio} (including an argument from each age group). However, it does not consider the order in which these arguments are presented, meaning the majority group will likely always be shown as the top argument. It also does not evaluate the distribution of the remaining arguments (after all values are covered). 

The fairness metric and the results for the representation analysis assess whether each group is represented in the top arguments according to its overall proportion. Nonetheless, there is a debate on whether this is fair because the majority group will be more frequently represented. An alternative approach would aim for an equal distribution of each group among the top arguments, ensuring that minority groups are as prominently represented as possible.

\paragraph{Data License}
All the data provided for this shared task is licensed under CC BY-NC 4.0, and the copyright remains with SmartVote (www.smartvote.ch).

\section*{Ethical Considerations}

\paragraph{Intended Use}
LMs have the potential to support the formation of opinions and foster a thorough and fine-grained discourse by navigating the diversity and large size of available political statements and standpoints.
While the data we use in this shared task are crucial for a comprehensive evaluation of LM's abilities regarding such supportive use cases, they have the potential for building manipulative systems. 
To ensure the data's supportive intent in this shared task, we will make it available solely upon request for research purposes and require concrete information about the specific usage.

\paragraph{Data Privacy}
For this shared task, we conducted an annotation study and collected personal information (demographic and socio-cultural variables) about the annotators.
As part of the obtained ethical clearance, we collected the explicit consent of the annotators during participation and relied on anonymized identifiers throughout the study. 
Therefore, we do not have any information about the specific person beyond the collected data.
Furthermore, we categorize more sensible information, like age, into different bins. 

Concerning the data provided by SmartVote (including the text of the arguments and the corresponding \textit{socio} profile of the politicians), we follow their privacy statement\footnote{Available \href{https://web.archive.org/web/20240726155426/https://www.smartvote.ch/en/wiki/methodology-questionnaire?locale=en_CH}{online}.}.
Specifically, the politicians agreed that all available public data on the platform could be shared anonymized.

\paragraph{Personalization}
Personalized recommendations of arguments based on one \textit{socio} are oversimplified and reduce diversity. The presented shared task started with a simplified scenario where only one socio was presented at a time since it was the first shared task. Given the rich and diverse profiles of authors and readers available, we advocate for more research on intersectionality and a broader, more nuanced representation of users in personalization research. 

As we have observed, there is a significant dataset bias with specific groups being underrepresented. Despite our efforts to incorporate diversity in the presented arguments for this shared task, this bias heavily influences systems. We advocate for further research and development of methods to diversify recommendations effectively. We see potential in combining personalization with diversification. For instance, while users tend to prefer arguments that align with their political attitudes, a system could optimize for this preference while presenting a range of perspectives, including arguments from different genders, age groups, and educational backgrounds. This approach would ensure a more pluralistic presentation of viewpoints while still showing arguments the user perceives as convincing or relevant.

\bibliography{anthology, anthology_p2, custom}
\clearpage
\appendix
\onecolumn
\section{Appendix}\label{sec:appendix}

\subsection{Details of the Annotation Study}\label{app:annotation_study}
Within the conducted annotation study, 22 annotators were asked to select intuitively relevant arguments for 27 political questions. 
Specifically, we conduct a two-staged study. 
First, collect the \textit{socio} variables from the annotators themselves using a survey to collect \textit{gender}, \textit{age}, \textit{civil status}, and \textit{denomination}.
Note, we remove \textit{residence} as a minority of the people were willing to share where they live.
Additionally, we collect their \textit{political attitude} and \textit{important political issues} using the same SmartVote questionnaire as filled out by the politicians. 
Secondly, we present 20 arguments for every 27 political questions and let the annotators choose those that intuitively address the given question from their perspective.

\begin{figure*}[t]
    \centering
    \includegraphics[width=0.95\textwidth]{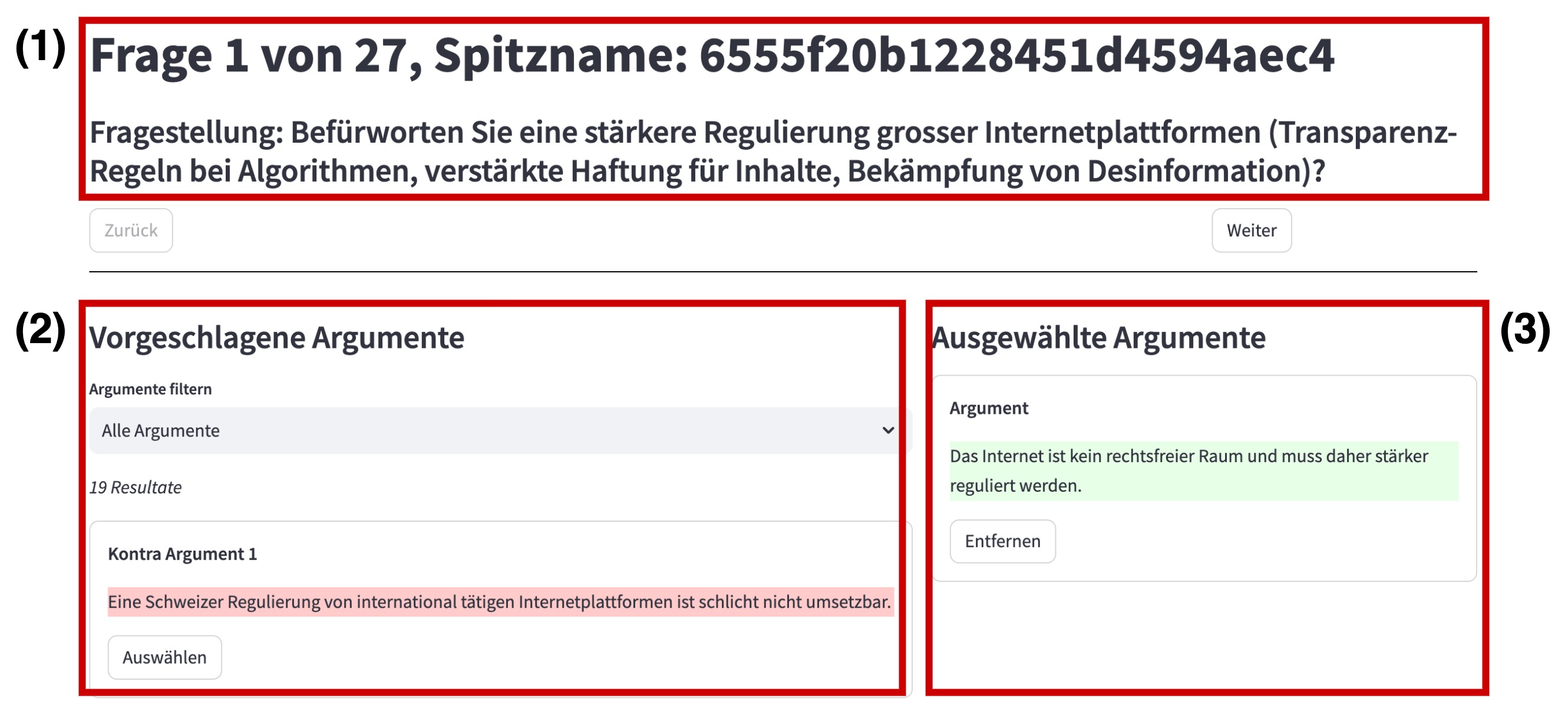}
    \caption{Screenshot of the annotation UI. It presents the annotator with the specific political question (1), 20 arguments addressing this question and allows to select the intuitively relevant ones (2), and list the already selected arguments (3).}
    \label{fig:annotation-ui}
\end{figure*}

\paragraph{Annotation Interface} 
We show an overview of the annotation UI in \autoref{fig:annotation-ui}. 
This interface presents the annotators one political question at a time, along with 20 arguments addressing this question from different perspectives.
Afterward, we ask the annotators to select which of the present arguments is intuitively relevant to them.  
Selected arguments will be listed on the right and can also be deleted later on.

\paragraph{Ethical Considerations}
As we collected demographic and socio-cultural data of the annotators, we collected the explicit consent of the annotators during the study. 
We inform them that we only collect categorized data, like the binned age, and that they can ask to delete it. 
This procedure has been approved by the ethical board of TU Darmstadt.
However, during preliminary discussion, it was decided that full ethical approval is unnecessary.

\paragraph{Payment}

We recruit the annotators on prolific and pay them an hourly rate of 25 Swiss francs. 
While there is no minimum wage in Switzerland, this salary is above the minimum.

\begin{figure*}[t]
    \centering
    \includegraphics[width=0.95\textwidth]{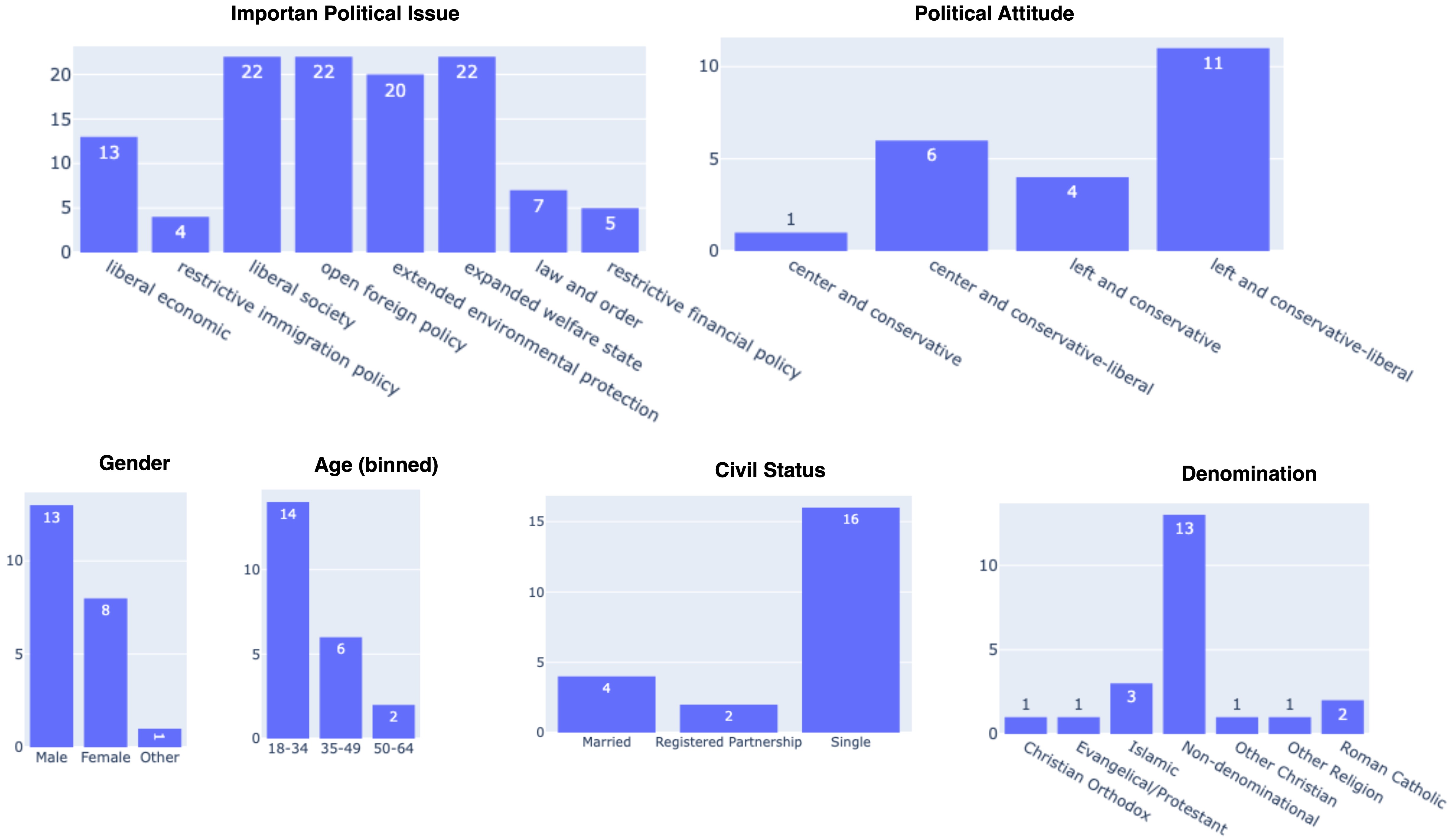}
    \caption{Distribution of the annotators' different demographic and socio-cultural variables: important political issues, political attitude, gender, age (binned), civil status, and denomination.
    Note that one person can have more than one important political issue.}
    \label{fig:socios_annotation_study}
\end{figure*}

\begin{figure*}
    \centering
    \includegraphics[width=0.95\textwidth]{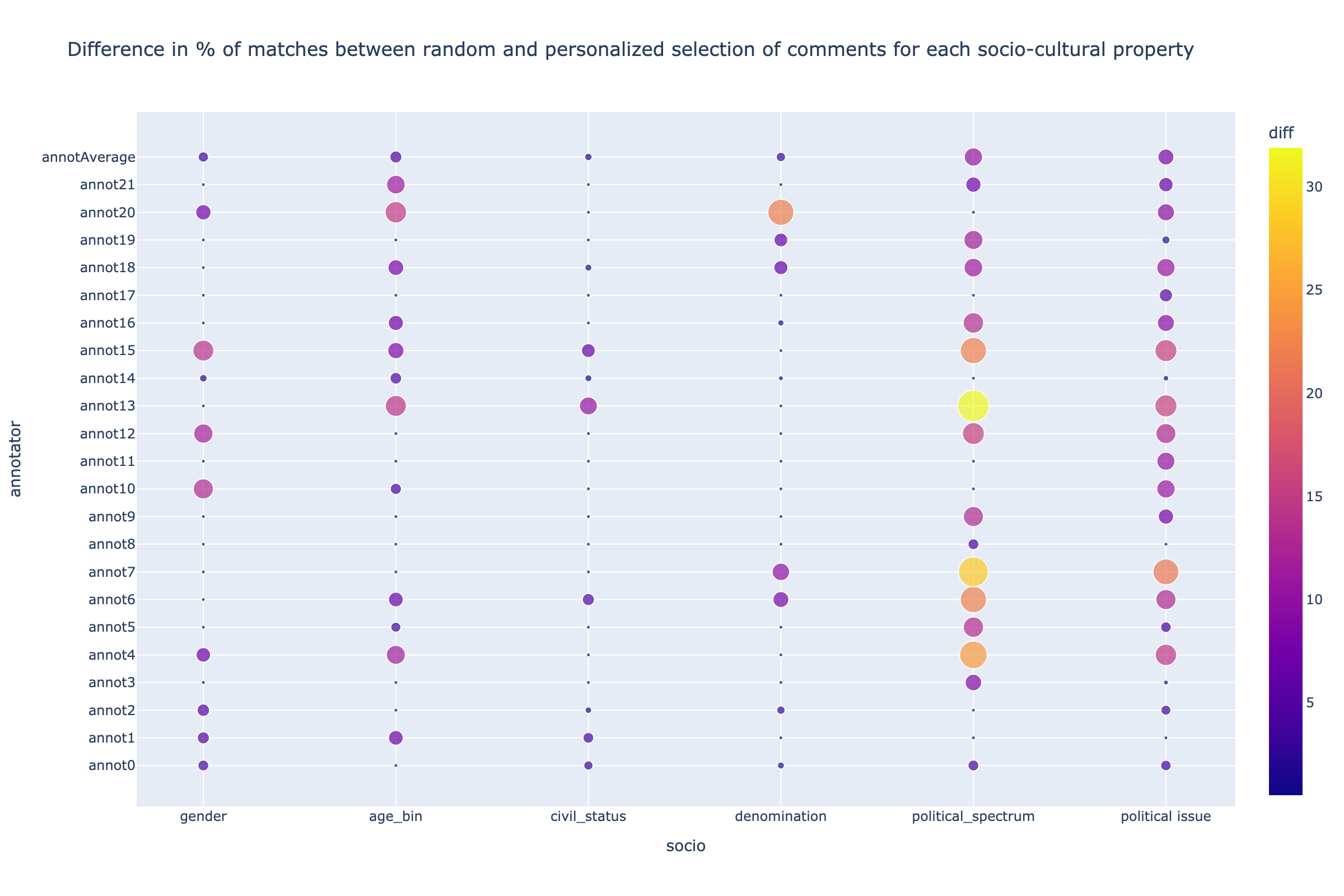}
    \caption{Amount of personalization per demographic and socio-cultural variable in the user study: percentage indicates the difference in matched arguments for a specific property when a user selects relevant arguments versus a random sample of relevant arguments.}
    \label{fig:personalization}
\end{figure*}

\paragraph{\textit{Socio} Variable of the Annotators}
We show in \autoref{fig:socios_annotation_study} the demographic and socio-cultural variables of the 22 annotators. 
However, the distribution is similar to the politicians' distribution (\autoref{sec:data}) but on a much smaller scale.
As a result, distinct values of a single variable are not covered. 
For example, we cover only four out of nine distinct political spectra. 
Further, we analyze in \autoref{fig:personalization} the agreement (personalization) of the annotator's perspectives with those of the authors whose arguments the annotator selects.
We found that annotators highly match with the authors' perspective regarding \textit{political spectrum} and \textit{important political issue}, and moderately \textit{age} and \textit{gender}.

\subsection{Detailed Results of Shared Tasks}\label{subsec:detailed-results}
\autoref{leaderboards-1}, \autoref{leaderboards-2}, and \autoref{leaderboards-3} show the detailed leaderboards for scenarios one, two, and three. 
When looking at the detailed results (per dataset and scenario), we find that no solution fits all: sometimes a team achieves a better score on one dataset (e.g., team sövereign outperforms the other teams on the dataset of the 2019 election, but not on the 2023 / user study dataset). 
This can be attributed to the fact that the LLM re-ranking is less effective at ranking arguments it has not seen before, whereas the 2019 data may have been included in its training data in some form.
The perspectivism scenarios are significantly more challenging than retrieving relevant arguments per topic (no perspectivism), particularly when the perspective is only implicitly encoded in the argument. 
This gap in performance highlights the need for further research on this issue, as perspectivist argument retrieval appears to be a particularly difficult problem. 
However, it is encouraging that most teams can outperform the baseline on these scenarios by a substantial margin. 
Their approaches to handling perspectivism are moving in the right direction.

\begin{table*}[]
\small
\resizebox{0.95\textwidth}{!}{%
\begin{tabular}{@{}lrrlrrlrr@{}}
\toprule
\multicolumn{3}{c}{2019}                      & \multicolumn{3}{c}{2023}                      & \multicolumn{3}{c}{user study}                \\ \midrule
\textbf{team}   & \textbf{rel} & \textbf{div} & \textbf{team}   & \textbf{rel} & \textbf{div} & \textbf{team}   & \textbf{rel} & \textbf{div} \\
sövereign       & 99.9        & 9.22        & twente-bms-nlp  & 93.6        & 87.0        & twente-bms-nlp  & 94.4        & 88.0        \\
GESIS-DSM       & 98.7        & 91.6        & sougata         & 92.0        & 85.5        & sougata         & 76.1        & 71.2        \\
sbert\_baseline & 98.6        & 91.6        & sövereign       & 89.5        & 82.7        & boulderNLP      & 75.7        & 70.3        \\
boulderNLP      & 98.6        & 91.3        & boulderNLP      & 88.5        & 82.2        & sövereign       & 63.7        & 59.5        \\
twente-bms-nlp  & 97.9        & 91.0        & sbert\_baseline & 85.5        & 79.3        & sbert\_baseline & 63.7        & 59.3        \\
sougata         & 97.9        & 90.5        & GESIS-DSM       & 85.5        & 79.3        & GESIS-DSM       & 62.8        & 59.2        \\
team031         & 90.4        & 84.4        & team031         & 80.6        & 75.3        & team031         & 59.2        & 55.0        \\
bm25\_baseline  & 65.1        & 62.9        & bm25\_baseline  & 73.7        & 69.0        & bm25\_baseline  & 36.8        & 34.2        \\ \bottomrule
\end{tabular}}
\caption{Scenario 1: No Perspectivsm}
\label{leaderboards-1}
\end{table*}

\begin{table*}[]
\small
\resizebox{0.95\textwidth}{!}{%
\begin{tabular}{@{}lllllllll@{}}
\toprule
\multicolumn{3}{c}{2019}                                                              & \multicolumn{3}{c}{2023}                                                              & \multicolumn{3}{c}{user study}                                                        \\ \midrule
\textbf{team}   & \multicolumn{1}{r}{\textbf{rel}} & \multicolumn{1}{r}{\textbf{div}} & \textbf{team}   & \multicolumn{1}{r}{\textbf{rel}} & \multicolumn{1}{r}{\textbf{div}} & \textbf{team}   & \multicolumn{1}{r}{\textbf{rel}} & \multicolumn{1}{r}{\textbf{div}} \\
twente-bms-nlp  & 89.5                            & 85.2                            & sövereign       & 82.3                            & 79.4                            & twente-bms-nlp  & 79.8                            & 79.3                            \\
sövereign       & 87.8                            & 84.4                            & twente-bms-nlp  & 79.8                            & 77.1                            & sövereign       & 67.3                            & 67.5                            \\
GESIS-DSM       & 83.5                            & 80.7                            & GESIS-DSM       & 72.2                            & 70.1                            & sougata         & 648                            & 65.9                            \\
sougata         & 68.4                            & 66.5                            & sougata         & 67.4                            & 66.3                            & GESIS-DSM       & 61.6                            & 62.9                            \\
sbert\_baseline & 22.2                            & 20.8                            & sbert\_baseline & 14.8                            & 14.2                            & team031         & 41.3                            & 40.1                            \\
team031         & 18.1                            & 17.2                            & team031         & 13.2                            & 12.5                            & sbert\_baseline & 40.6                            & 40.0    \\ \bottomrule                
\end{tabular}}
\caption{Scenario 2: Explicit Perspectivsm}
\label{leaderboards-2}
\end{table*}

\begin{table*}[]
\small
\resizebox{0.95\textwidth}{!}{%
\begin{tabular}{@{}lllllllll@{}}
\toprule
\multicolumn{3}{c}{2019}                                                              & \multicolumn{3}{c}{2023}                                                              & \multicolumn{3}{c}{user study}                                                        \\ \midrule
\textbf{team}   & \multicolumn{1}{r}{\textbf{rel}} & \multicolumn{1}{r}{\textbf{div}} & \textbf{team}   & \multicolumn{1}{r}{\textbf{rel}} & \multicolumn{1}{r}{\textbf{div}} & \textbf{team}   & \multicolumn{1}{r}{\textbf{rel}} & \multicolumn{1}{r}{\textbf{div}} \\
sövereign       & 21.3                            & 19.9                            & twente-bms-nlp  & 14.9                            & 14.3                            & twente-bms-nlp  & 65.5                            & 63.6                            \\
twente-bms-nlp  & 20.3                            & 19.0                            & sövereign       & 13.9                            & 13.2                            & GESIS-DSM       & 471                            & 45.4                            \\
sbert\_baseline & 20.2                            & 18.9                            & GESIS-DSM       & 13.9                            & 13.2                            & sövereign       & 43.6                            & 42.5                            \\
GESIS-DSM       & 20.2                            & 18.9                            & sbert\_baseline & 13.6                            & 13.1                            & team031         & 41.3                            & 40.1                            \\
team031         & 18.1                            & 17.2                            & team031         & 13.2                            & 12.5                            & sbert\_baseline & 40.9                            & 39.7   \\ \bottomrule                                  
\end{tabular}}
\caption{Scenario 3: Implicit Perspectivsm}
\label{leaderboards-3}
\end{table*}

\subsection{Results regarding different top-k}\label{app:subsec_top-k}
\begin{figure}[t]
    \centering
    \includegraphics[width=0.48\textwidth]{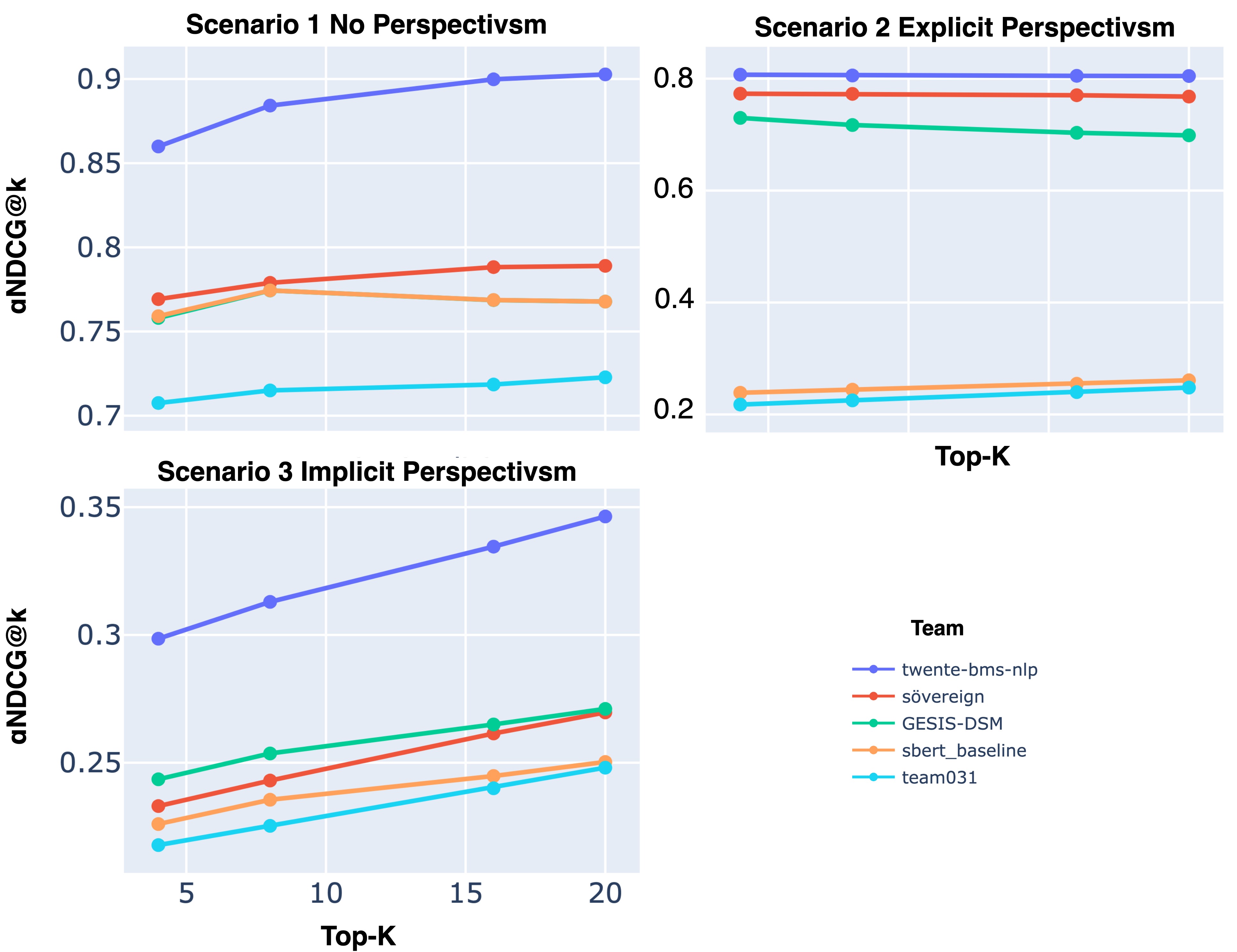}
    \caption{Overview of the per team performance regarding diversity (y-axis, $\alpha$NDCG@k) regarding top\-4, top\-8, top\-16, and top\-20 retrieved candidates for the three scenarios.}
    \label{fig:top-k}
\end{figure}

We analyze how the number of retrieved candidate arguments affects the performance.
From \autoref{fig:top-k}, the performance decreases with a higher $k$ for the first scenario (\textit{no perspectivism}) for the baseline and the submissions. 
Interestingly, this effect is less pronounced for the second scenario (\textit{explicit perspectivism}) and reversed for the third one (\textit{implicit perspectivism}). 
Specifically, three teams (\texttt{twente-bms-nlp}, \texttt{sövereign}, and \texttt{team031}) show more improvements with higher $k$ than the other teams.
These patterns indicate that their filtering or argument re-ranking methods work better on higher $k$.

\subsection{Analysis of bias in representation and performance}
\begin{figure}[htbp]
    \centering
    \begin{minipage}[b]{0.45\textwidth}
        \centering
        \includegraphics[width=\textwidth]{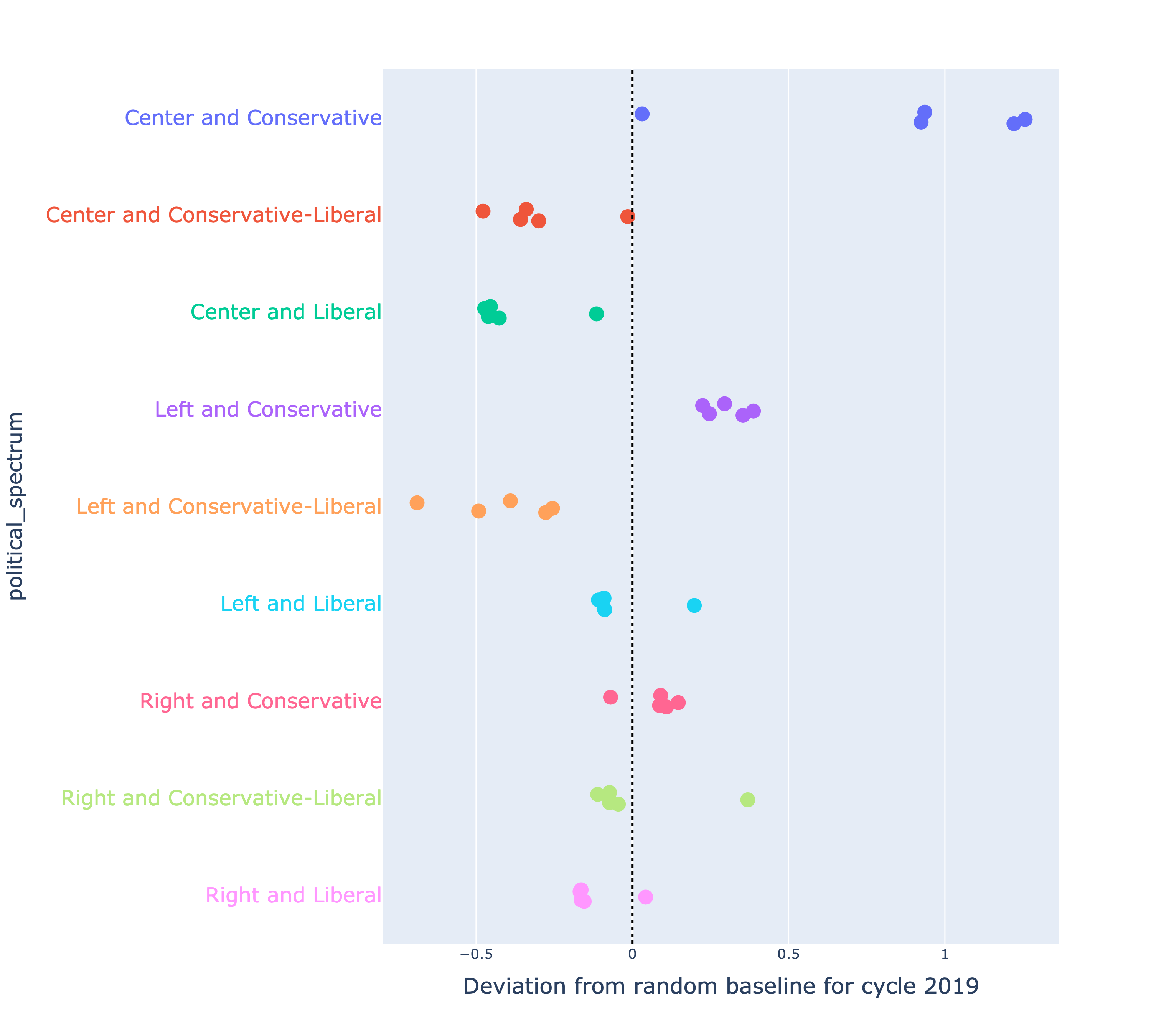}
        \label{fig:figure1}
    \end{minipage}
    
    \vspace{-5mm}
    \begin{minipage}[b]{0.45\textwidth}
        \centering
        \includegraphics[width=\textwidth]{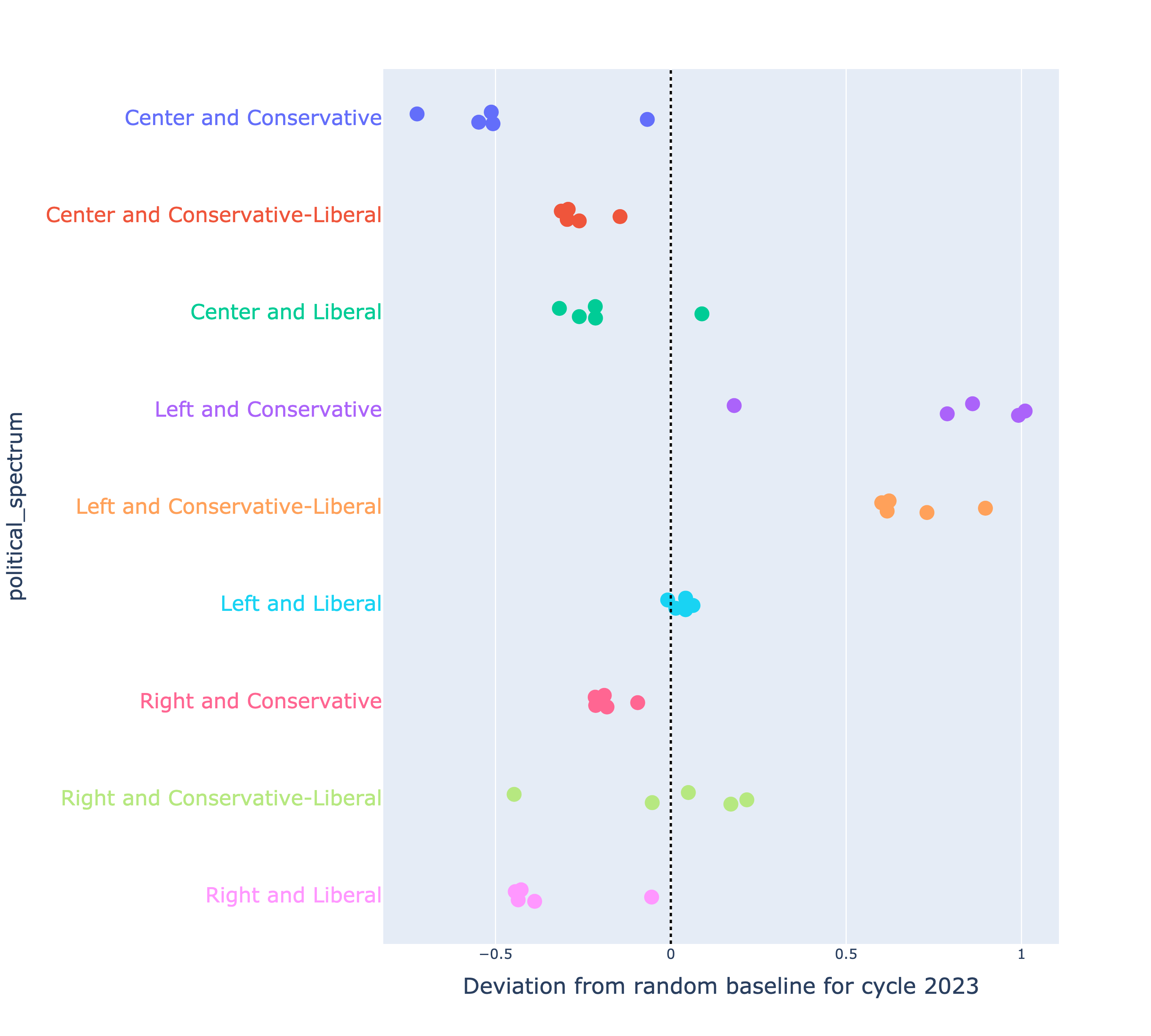}
        \label{fig:figure2}
    \end{minipage}
    
    \vspace{-3mm}
    \caption{Extent of system deviation from random sampling representing each political spectrum among the 20 most relevant arguments.}
    \label{fig:politicalbias}
\end{figure}

\autoref{fig:politicalbias} shows the representation bias of the different systems in representing different \textbf{political orientations}. We can observe a shift from 2019 to 2023 in representing the center/conservative group (over- then underrepresented), which can be accounted for the shift in topics. In both years we can observe that the data bias for left and conservative is reinforced, for left and conservative-liberal its reduced in 2019 but reinforced in 2023. 

\begin{figure}[htbp]
    \centering
    \begin{minipage}[b]{0.45\textwidth}
        \centering
        \includegraphics[width=\textwidth]{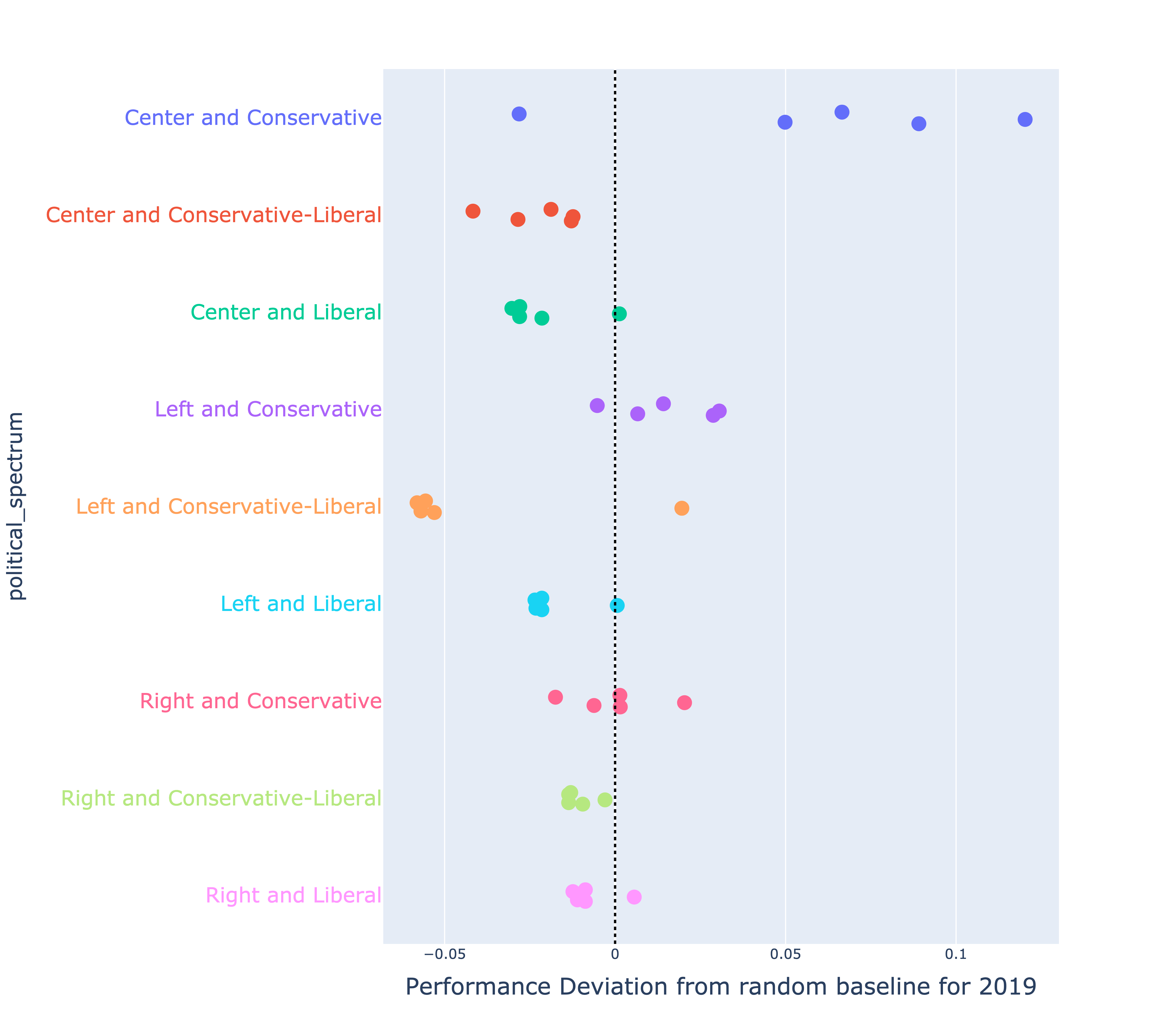}
        \label{fig:figure1}
    \end{minipage}

     \vspace{-5mm}
    \begin{minipage}[b]{0.45\textwidth}
        \centering
        \includegraphics[width=\textwidth]{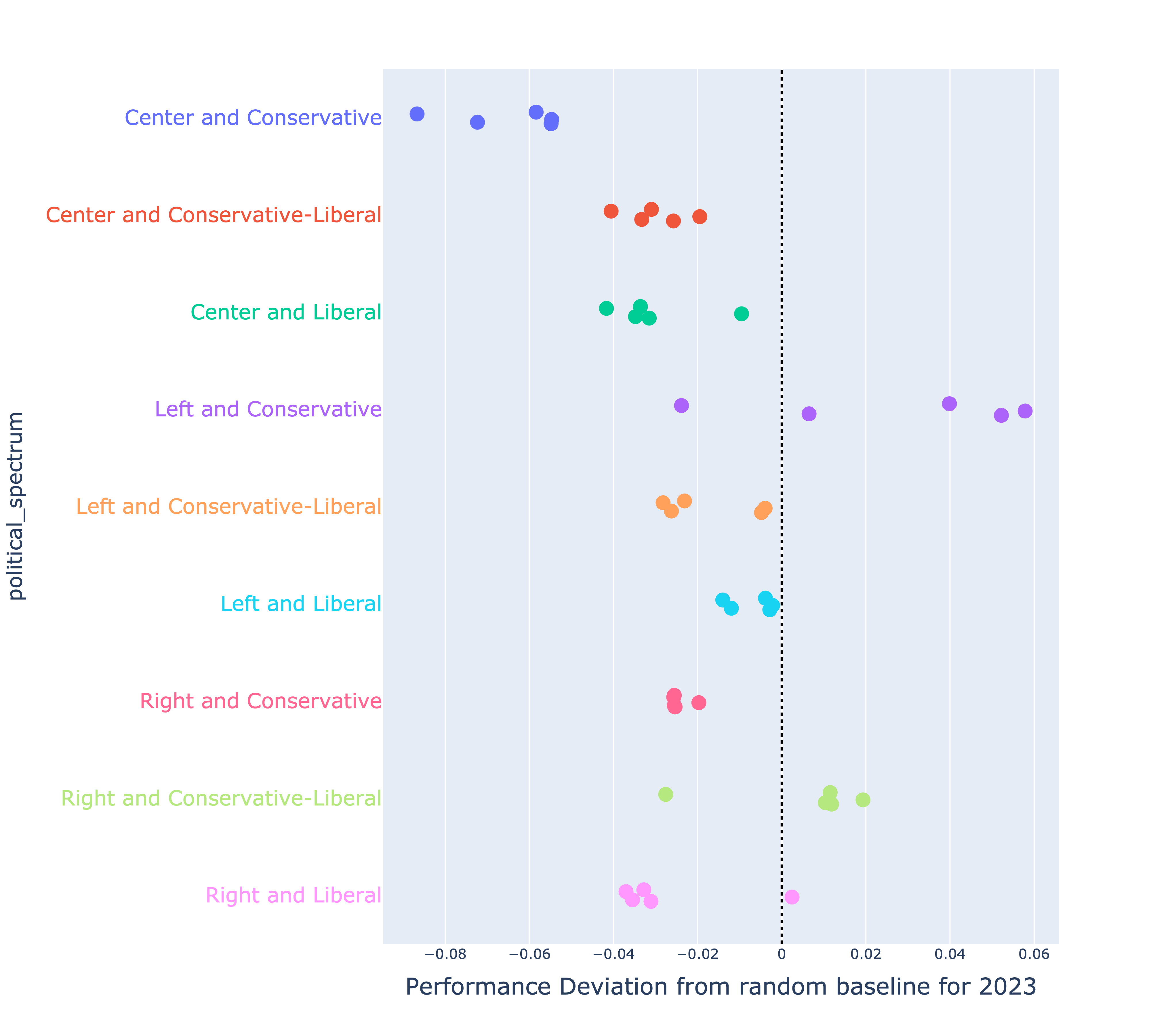}
        \label{fig:figure2}
    \end{minipage}

     \vspace{-3mm}
    \caption{Extent of system deviation from random sampling in performance from the nDCG score for different groups of political spectrum.}
    \label{fig:politicalperformance}
\end{figure}

\autoref{fig:politicalperformance} shows that some systems reduce and some reinforce the bias for left-(conservative/liberal) political orientation as the performance increases or decreases for those groups compared to the baseline.

\begin{figure}[htbp]
    \centering
    \begin{minipage}[b]{0.45\textwidth}
        \centering
        \includegraphics[width=\textwidth]{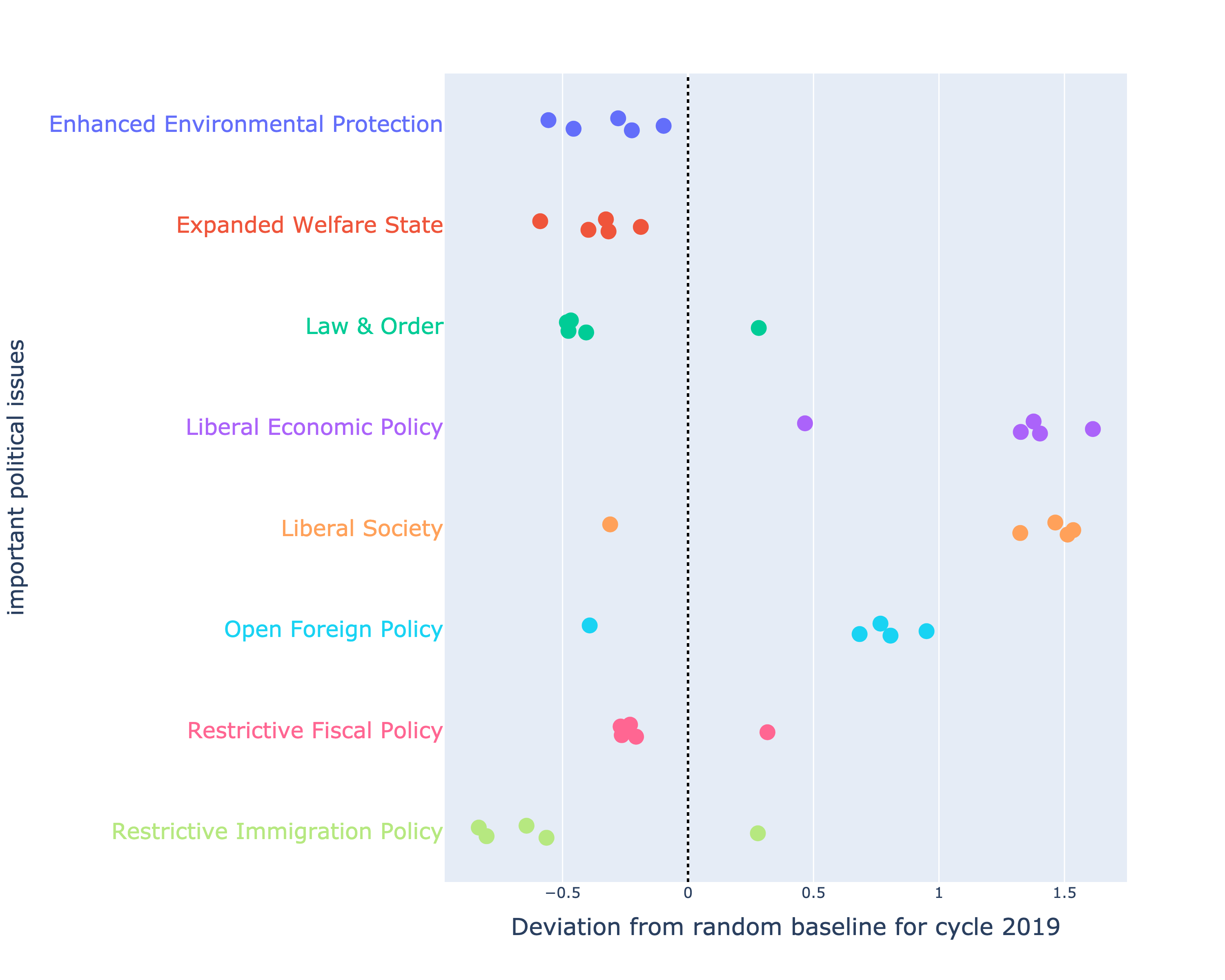}
        \label{fig:figure1}
    \end{minipage}
    
    \vspace{-5mm}
    \begin{minipage}[b]{0.45\textwidth}
        \centering
        \includegraphics[width=\textwidth]{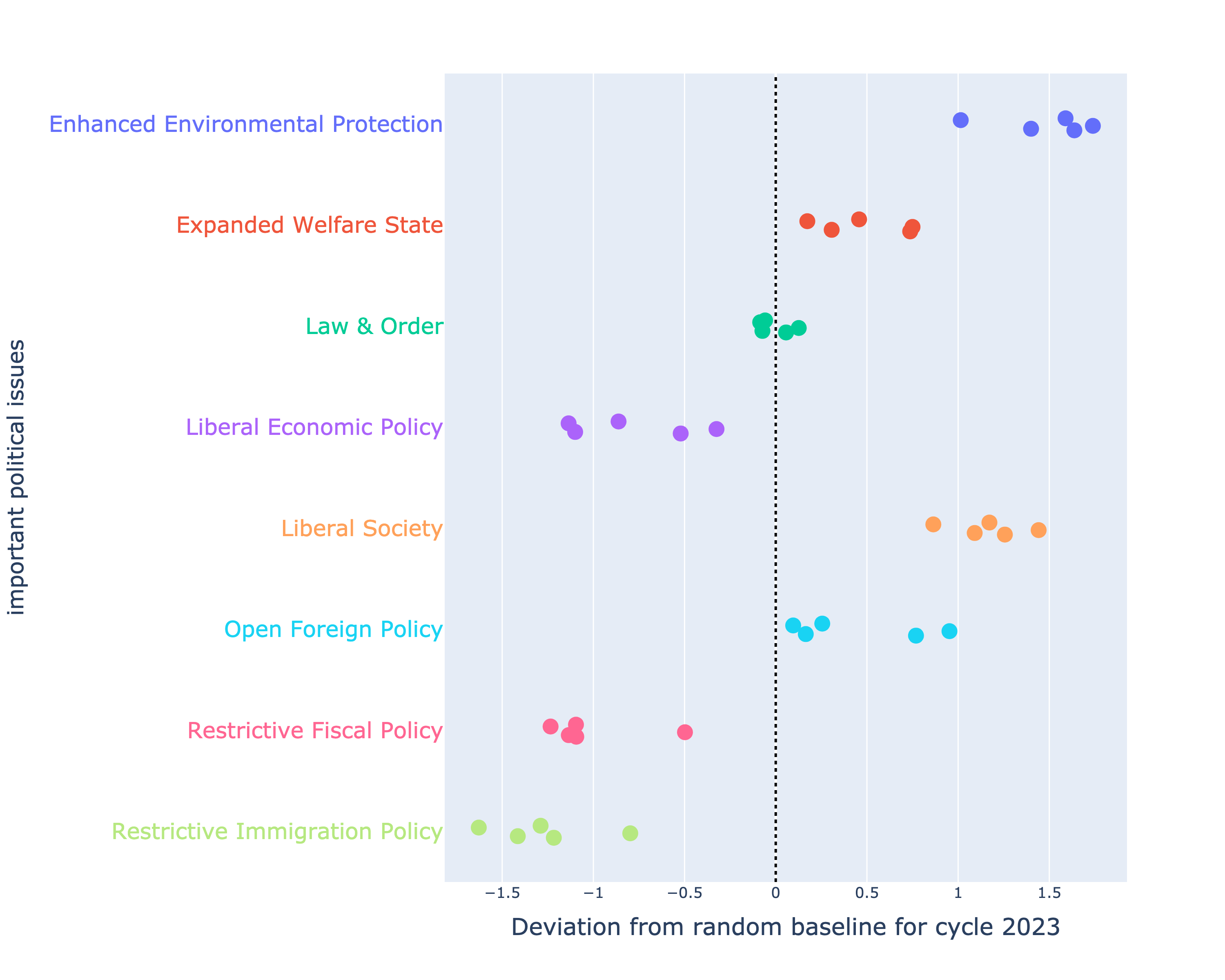}
        \label{fig:figure2}
    \end{minipage}
    
    \vspace{-3mm}
    \caption{Extent of system deviation from random sampling representing each important political issue among the 20 most relevant arguments.}
    \label{fig:issuebias}
\end{figure}

\begin{figure}[htbp]
    \centering
    \begin{minipage}[b]{0.45\textwidth}
        \centering
        \includegraphics[width=\textwidth]{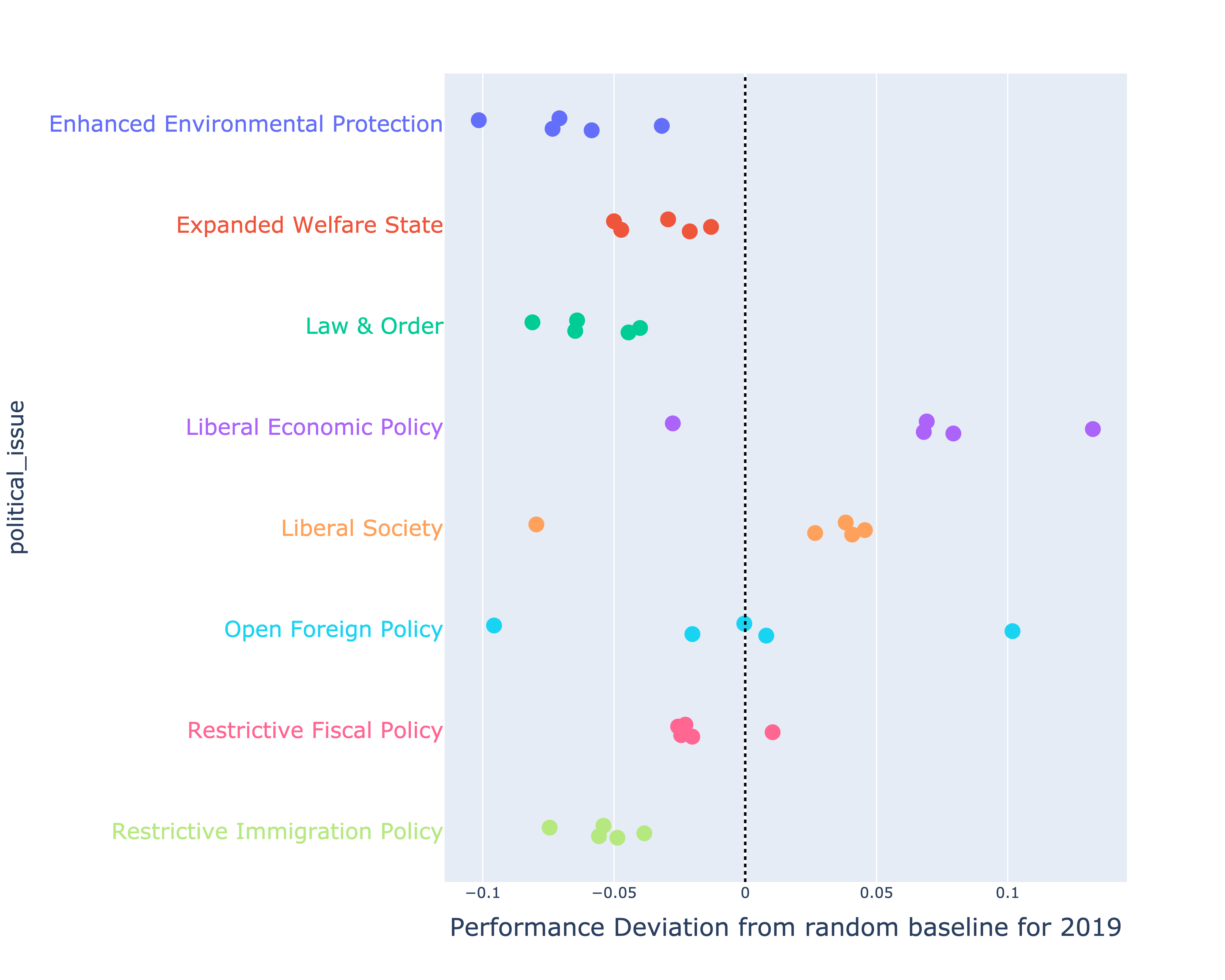}
        \label{fig:figure1}
    \end{minipage}
     \vspace{-3mm}
    \caption{Extent of system deviation from random sampling in performance from the nDCG score for different groups of political spectrum.}
    \label{fig:issueperformance}
\end{figure}

\autoref{fig:issuebias} shows a lot of diversity in teams when looking at the representation of \textbf{important political issues} compared to the other socio-cultural properties which can be accounted to the strong semantic influence they have on the text, i.e. it is likely that an important political issue is expressed in the framing of the argument. This is especially the case for the election of 2019, since this data was used for training the systems, and some classifiers were used to predict which issues were important for an author of a certain argument. Some teams retrieve more arguments for law and order, liberal society, or open foreign policy, while others retrieve significantly fewer than the baselineb.
This only partially impacts the results  (\autoref{fig:issueperformance}), e.g., for law and order, all systems underperform, and over-representing open foreign policy does not increase the performance of all systems on that issue.

\begin{figure}[htbp]
    \centering
    \begin{minipage}[b]{0.45\textwidth}
        \centering
        \includegraphics[width=\textwidth]{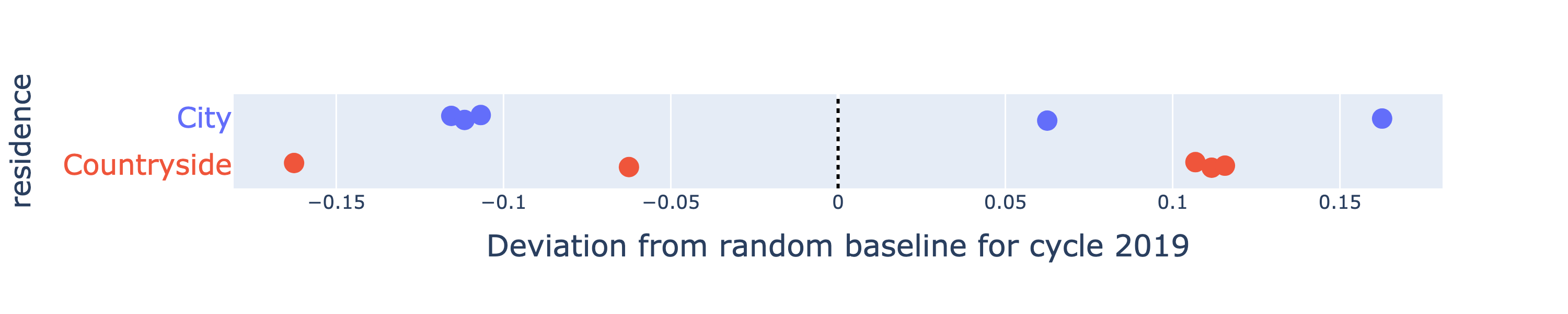}
        \label{fig:figure1}
    \end{minipage}
    
    \vspace{-5mm}
    \begin{minipage}[b]{0.45\textwidth}
        \centering
        \includegraphics[width=\textwidth]{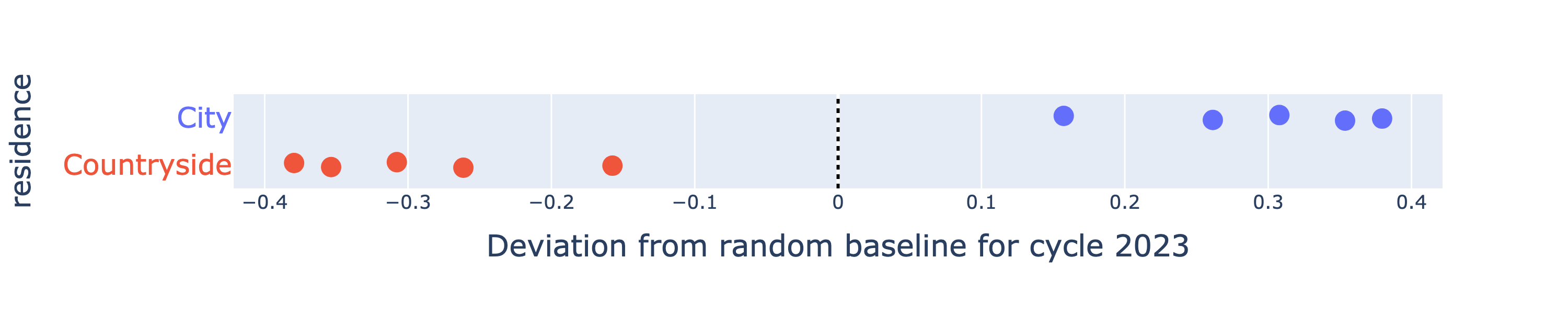}
        \label{fig:figure2}
    \end{minipage}
    
    \vspace{-3mm}
    \caption{Extent of system deviation from random sampling representing each important residence group among the 20 most relevant arguments.}
    \label{fig:residencebias}
\end{figure}

For \textbf{residence} we find significant differences between the elections: systems are split between reinforcing or reducing the bias of arguments by authors from countryside in 2019, in 2023 all systems reduce that bias (\autoref{fig:residencebias}). This weakly impacts performance, slightly mitigating the countryside bias for a few systems in 2019 and gaining small improvements for arguments from authors from the city in 2023. 

\begin{figure}[htbp]
    \centering
    \begin{minipage}[b]{0.45\textwidth}
        \centering
        \includegraphics[width=\textwidth]{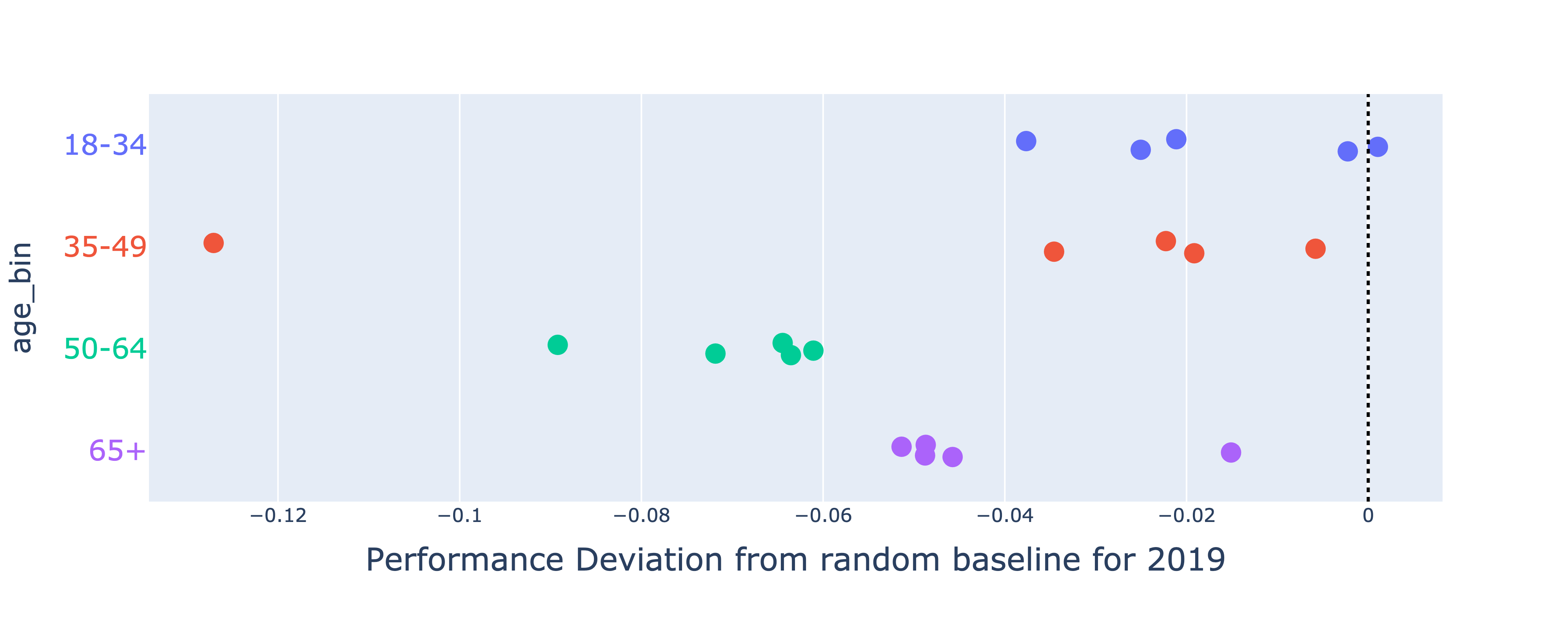}
        \label{fig:figure1}
    \end{minipage}

     \vspace{-5mm}
    \begin{minipage}[b]{0.45\textwidth}
        \centering
        \includegraphics[width=\textwidth]{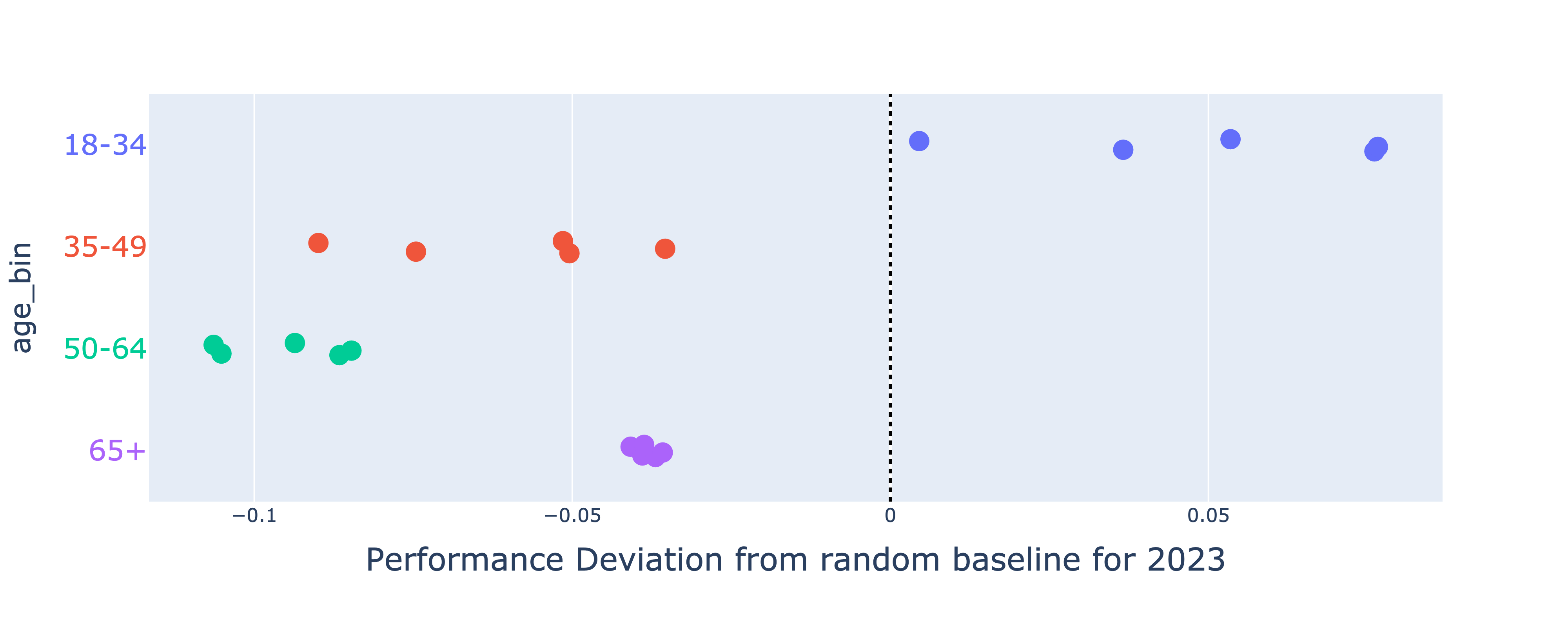}
        \label{fig:figure2}
    \end{minipage}

     \vspace{-3mm}
    \caption{Extent of system deviation from random sampling in performance from the nDCG score for different age groups.}
    \label{fig:ageperformance}
\end{figure}

\begin{figure}[htbp]
    \centering
    \begin{minipage}[b]{0.45\textwidth}
        \centering
        \includegraphics[width=\textwidth]{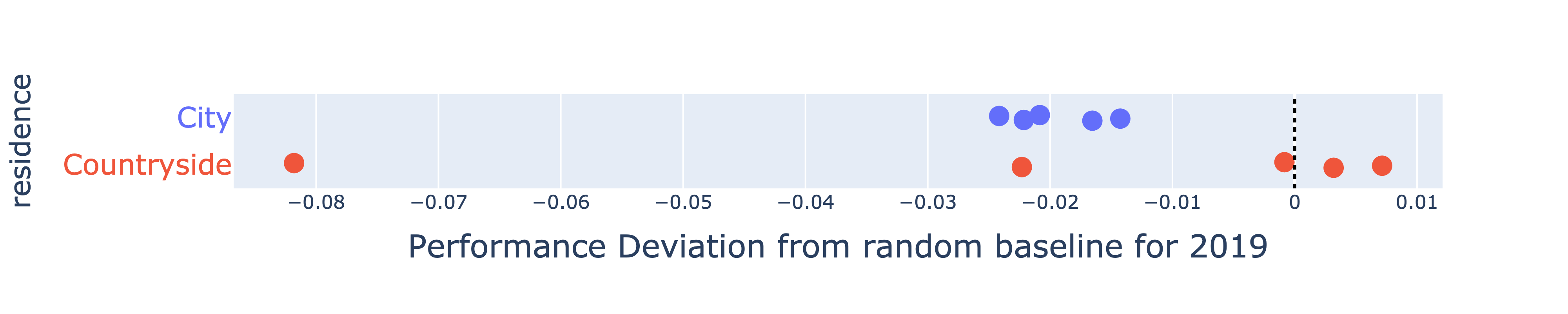}
        \label{fig:figure1}
    \end{minipage}

     \vspace{-5mm}
    \begin{minipage}[b]{0.45\textwidth}
        \centering
        \includegraphics[width=\textwidth]{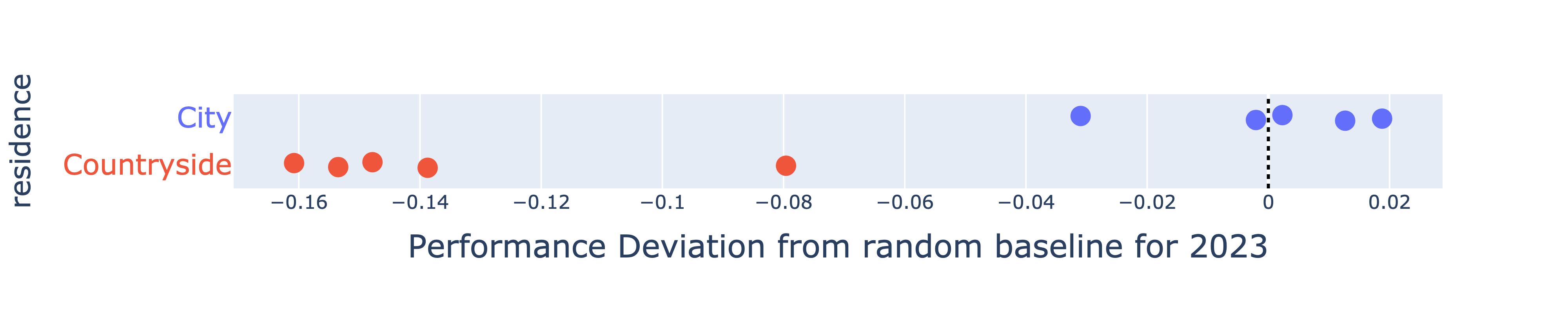}
        \label{fig:figure2}
    \end{minipage}

     \vspace{-3mm}
    \caption{Extent of system deviation from random sampling in performance from the nDCG score for residence.}
    \label{fig:residenceperformance}
\end{figure}
\end{document}